\documentclass[12pt]{uwthesis17}
\usepackage{amsmath, amssymb, amsthm, amsxtra}
\usepackage{float}
\usepackage{extramarks}
\usepackage{fancyhdr}
\usepackage{algorithm}
\usepackage{algpseudocode}
\usepackage{booktabs}
\usepackage{subcaption}
\usepackage[hidelinks]{hyperref}
\usepackage{booktabs}
\usepackage{bbm}

\newcommand*{\Z}{\mathbb{Z}}
\renewcommand\thesection{\arabic{section}}
\algdef{SE}[SUBALG]{Indent}{EndIndent}{}{\algorithmicend\ }%
\algtext*{Indent}
\algtext*{EndIndent}

\usepackage{tocloft}
\DeclareUnicodeCharacter{202F}{\,}

\usepackage{ifpdf}
\ifpdf 
  \usepackage[pdftex]{graphicx}
\else 
  \usepackage[dvips]{graphicx}
\fi
\title{Validation and Inference of Agent Based Models}
\author{Dale Townsend}
\begin{document}
\pagenumbering{roman}
\maketitle
\setcounter{page}{2}
\begin{abstract}
Agent Based Modelling (ABM) is a computational framework for simulating the behaviours and interactions of autonomous agents. As Agent Based Models are usually representative of complex systems, obtaining a likelihood function of the model parameters is nearly always intractable. There is a necessity to conduct inference in a likelihood free context in order to understand the model output. Approximate Bayesian Computation is a suitable approach for this inference. It can be applied to an Agent Based Model to both validate the simulation and infer a set of parameters to describe the model. Recent research in ABC has yielded increasingly efficient algorithms for calculating the approximate likelihood. These are investigated and compared using a pedestrian model in the Hamilton CBD.
\end{abstract}
\begin{acknowledgements}
I would like to thank Chaitanya Joshi for guidance in producing this dissertation. His expertise and motivation was instrumental throughout my research. \\

I am grateful to Hamilton City Council for allowing me to develop the CBD pedestrian model as a demonstration of my research. The input from staff members in the City Transportation Unit was invaluable in guiding the output and use cases for the model. Particular thanks to John Kinghorn, Simon Crowther, and Andrea Timings. \\

A special thanks to Charn\'e Nel, who was always there for support (and coffee) while we were both working on our respective dissertations.
\end{acknowledgements}
\newpage
\topskip0pt
\vspace*{\fill}
\textit{Dedicated to my grandfather, Gordon Townsend (1931-2018).} \\
\vspace*{\fill}
\newpage
\tableofcontents
\newpage
\listoffigures
\newpage
\listoftables
\newpage
\section{Glossary}
\textbf{ABM} - Agent Based Modelling \\
\textbf{ABC} - Approximate Bayesian Computation \\
\textbf{ODD} - Overview, Design, Details \\
\textbf{BDI} - Belief-Desire-Intention \\
\textbf{HCC} - Hamilton City Council \\
\textbf{CBD} - Central Business District in Hamilton, New Zealand \\
\textbf{AW} - Count data collected for the southwest direction at the Anglesea-Ward intersection \\
\textbf{TR} - Count data collected for the northeast direction on Ward St towards Victoria St \\
\textbf{TA} - Count data collected for the southeast direction on Worley Pl \\
\textbf{CPS} - Count data collected for the southeast direction out of Centre Place (South) to Civic Square \\
\newpage
\pagenumbering{arabic}
\setcounter{page}{1}

\section{Introduction}

Agent Based Modelling refers to a class of computational models invoking the dynamic behaviour, reactions and interaction amongst a set of agents in a shared environment (Abar, Theodoropoulos, Lemarinier, O'Hare, 2017). A wide range of real world systems can be  modelled using this framework. It is particularly effective in complex systems of multiple heterogeneous, interacting agents. Valuable information can be gathered by observing the aggregate behaviour of a micro level simulation where agents interact with both other agents and the environment. \\

A key component of ABM research is in the post analysis of the model output. Due to its stochastic simulation nature there is a vast flexibility in the quantity and type of data produced by a given ABM. As a result, it is necessary to output only the most relevant data and analyse it using efficient techniques for effectively answering the original set of research questions. In a traditional Bayesian context, there exists a set of random parameters describing the function of an ABM. These parameters are usually inferred by fitting a model to the data and calculating the probability of observing the parameters given the data (\textit{posterior distribution}). However, it is often the case that a complex system has an intractable likelihood, where the probability of the parameters given the data cannot be calculated. \\

In this setting, alternative approaches must be examined to find the posteiror distribution. One such `likelihood free' approach is Approximate Bayesian Computation (ABC). Where observed data can be compared against simulated data, ABC can be used to infer a set of parameters for a model. This turns out to be a suitable method for the analysis of an agent based model, where the likelihood is often intractable. Various examples of analysing an ABM using the ABC algorithm are shown, with the focus being on modelling pedestrians flows in the Hamilton CBD. ABC is used to validate the ABM by approximating the observed data, as well as infer a set of probability parameters for the movement of pedestrians.

\section{Agent Based Modelling}

Agent Based Modelling (ABM) is a computational framework for simulating the behaviours and interactions of autonomous agents. The agents can represent any individual or collective entity within a complex system. They each have a set of properties and behaviours, which are carried out at each time step of the model simulation. The output of the model gives us a means of assessing how the actions of individual entities affect the system as a whole. In a complex system it is often the case that an emergent pattern is seen, but the actions of the agents making up the system is not readily visible. Constructing a model using the ABM framework allows us to understand how the properties and behaviours of individual agents develop this emergent pattern. \\

Joslyn \& Rocha, (2000) define a complex system as ``consisting of a large number of interacting components (agents, processes, etc.) whose aggregate activity is non-linear (not derivable from the summations of the activity of individual components), and typically exhibits hierarchical self-organization under selective pressures". Such systems contain a number of complexities in state space, interactions, behaviour and spatio-temporal structure. A key property is that of emergence. When a large number of agents in a complex system interact in a spatio-temporal structure, properties can be observed which are usually not readily apparent from observing the agents independently. ABM allows us to construct a complex system in such a way that the aggregate behaviour of individual agents can be observed. By constructing such a model on a microscale level, it is often the case that more information can be extracted than viewing the model solely by its aggregate behaviour. \\

ABM has been used across a wide range of scientific disciplines, including ecology, economics, and biology. In the biological sciences, recent advances in computational tools have allowed for the simulation of individual cells in order to observe the aggregate behaviour. One application is the modelling of the chemotactic response of \textit{E.coli} cells to chemo attractant gradients in a 3D environment (Emonet et al, 2005). Representing cells as autonomous agents competing in a spatial environment is well suited for such biological analysis, where the physical observation of the process is either not possible or practical. More recently, ABM has been used in urban planning of city environments. As the dynamic structure of transportation modes evolves and the use of alternative modes increases, is it now vital to understand how traffic flows change across an urban environment. A tool developed by Aschwanden, Wullschleger, Müller, \& Schmitt (2012), allowed for the simulation of a city's transportation network to evaluate greenhouse gas emissions and pedestrian flow (Aschwanden et al, 2012).\\

Much of the phenomena observed in the world can be modelled with an ABM. For example consider the ecological predator-prey model. There are two popular approaches to understanding the dynamics of the change in populations between two species where one predates on the other. One is known as the \textit{Lotka--Volterra} model, developed independently by Lotka (1925) and Volterra (1926). This pair of first-order non linear differential equations describe the change in the population of two species over time.
\begin{align}
\frac{dx}{dt} = \alpha x - \beta x y \\
\frac{dy}{dt} = \delta x y - \gamma y,
\end{align}
where $x$ = the population of the prey species, \\
$y$ = the population of the predator species, \\
$t$ = time, \\
$\frac{dx}{dt}, \frac{dy}{dt}$ represent the instantaneous growth of the two populations, \\
and $\alpha, \beta, \gamma, \delta$ are positive real parameters which describe the interaction between the two species. Each parameter can be adjusted and a solution yielded to find the change in population of two interacting species. While the Lotka-Volterra model has a long history of use in ecology, it is too simplistic to apply to ecosystems where predator competition is involved. In addition, it is not possible for either species in the model to saturate completely. \\

An alternative method to describe predator-prey interaction is that of an agent based model. Consider the interaction between wolf and sheep in a field. First we describe this model by answering a set of key questions. Describing an agent based model in this way enables another researcher to quickly and easily understand its function.

\begin{enumerate}
\item{\textbf{Research Question}} \\
Under what conditions is an equilibrium of the wolf and sheep population levels obtained, when wolves predate on the sheep and sheep consume grass?

\item{\textbf{Agents}} \\
Wolves, Sheep, Grass

\item{\textbf{Agent Properties}} \\
Wolf and Sheep: Energy, Location, Heading. Grass: Amount of grass

\item{\textbf{Agent Behaviours}} \\
Wolf and Sheep: Move, Die, Reproduce. Wolf: Eat sheep. Sheep: Eat Grass. Grass: Grow

\item{\textbf{Parameters}} \\
Num Sheep, Num Wolves, Movement Cost, Energy Gain from grass, energy gain from sheep, grass regrowth rate

\item{\textbf{Time Step}} \\
Sheep \& Wolves: 1. Move, 2. Die, 3. Eat, 4. Reproduce; Grass: 5. Grow

\item{\textbf{Measures}} \\
Sheep and wolf population versus time
\end{enumerate}

The two species can be represented as two sets of agents. Each agent of the set is given relevant properties chosen as having the most influence on deriving the output we want to observe.  Both the sheep and wolves are given three shared properties: an energy level, location and heading (direction). There is no limit to the number of properties, but for simplicity purposes, they should be restricted to those necessary to answer the research question. The second step is to describe the environment in which the agents interact. Here the environment is a field with patches of grass. This could be extended to include fences, elevation and other features, but again for simplicity purposes are left out of the first model. When the agents reach the end of the world, they will reappear on the other side of the world. This is known as a `torus shaped' boundary condition, and is a common feature in ABMs. Next, the agent behaviour is described. Both sheep and wolves have the ability to move, eat, reproduce, and die. Where they differ is in what each species eat - wolves will consume sheep while sheep will consume grass. Parameters are then chosen to control various attributes of the model. This allows the observation of changes in the model given varying parameters - the numbers of each population, the amount of grass available, or the reproduction rate can be changed and the aggregate effect of each observed. At each time step (or \textit{tick}, which will be used hereafter), the agents carry out some action. Both the sheep and wolves move, die, eat then reproduce. The grass grows at each tick. The populations of the two species is observed as the aggregate behaviour.

\begin{figure}[H]
    \centering
    \includegraphics[width=10cm]{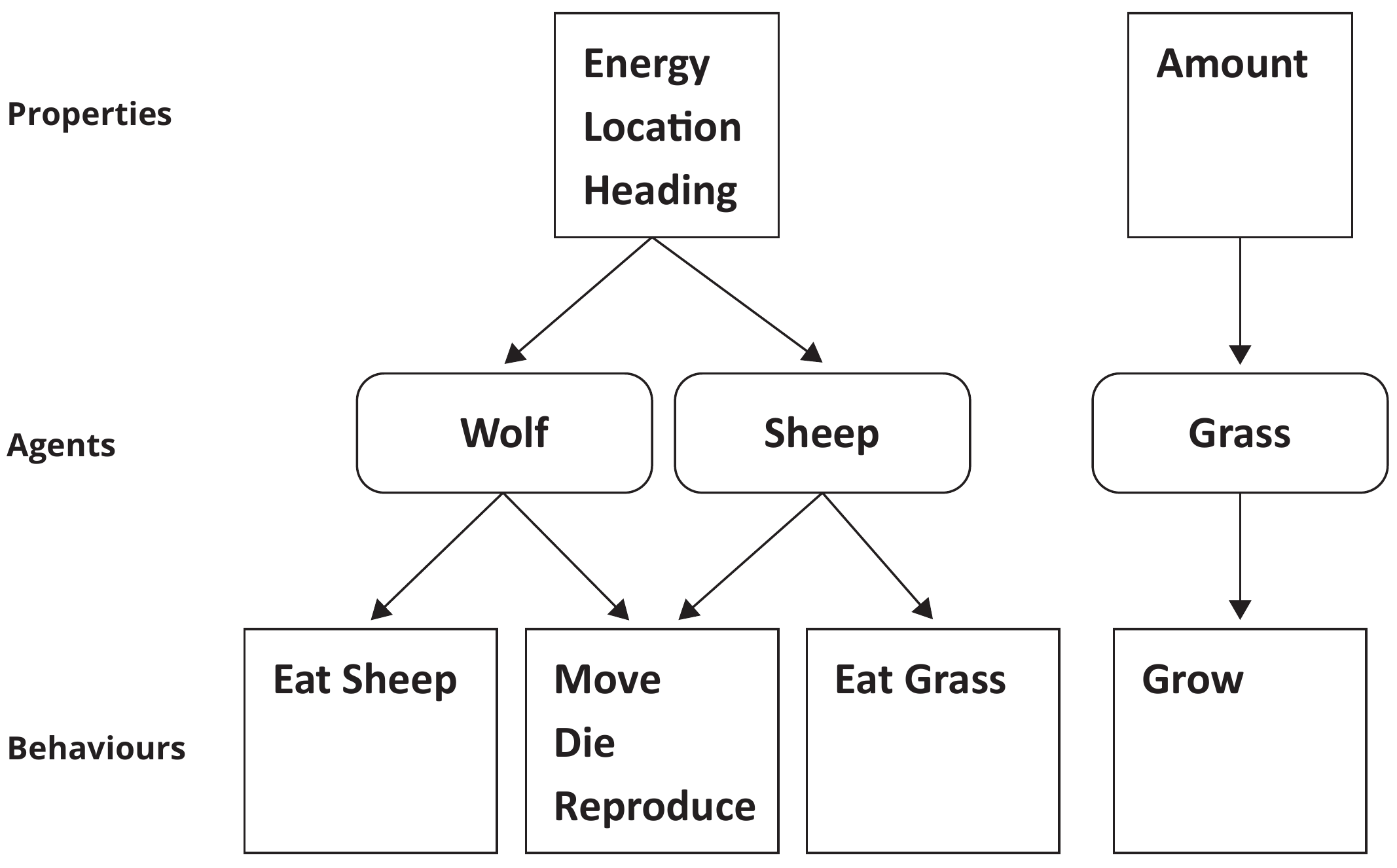}
    \caption{Properties, Agents and Behaviours in the Wolf-Sheep predation model}
\end{figure}

\subsection{Analysis of an ABM}

In the above model there exists a set of parameters which achieve equilibrium in the sheep and wolf populations. This is often found through trial and error. Another, more sophisticated method involves looping through many thousands of simulations, changing the set of parameters each time. The sheep and wolf populations are analysed in each run. The output is some metric describing how close the populations are to sustaining over time. This is demonstrated below. In each run, the parameters are adjusted, the model is run for a set number of time ticks, and the final numbers of both wolf and sheep compared to the initial populations. The data can be analysed with appropriate statistical models to identify stationarity. One such method is to view the two populations as two time series. The oscillation of the species populations can be viewed as seasonality. Given enough simulation time, the difference in population numbers from the start to the end of the time series is compared to determine if the system has reached equilibrium. \\

The BehaviorSpace tool in ABM software NetLogo allows a simulation to be run multiple times with changing parameters. A simple demonstration of this tool for analysis is when the `energy from grass' parameter is changed. This parameter determines the energy that sheep gather from eating grass. Because sheep lose energy from movement, a low amount of energy gained from grass is likely to result in a larger decrease in the sheep population than if a large amount of energy was gained from eating grass. The Wolf-Sheep predation ABM allows this theory to be tested in a simulation based manner, followed by the application of robust statistical analysis methods from the output data to answer questions about the model. Keeping all other parameters fixed, the grass energy parameter was updated. The simulation was run for 1000 ticks and the population of the two species recorded at each tick.

\begin{figure}[H]
    \centering
    \includegraphics[width=14cm]{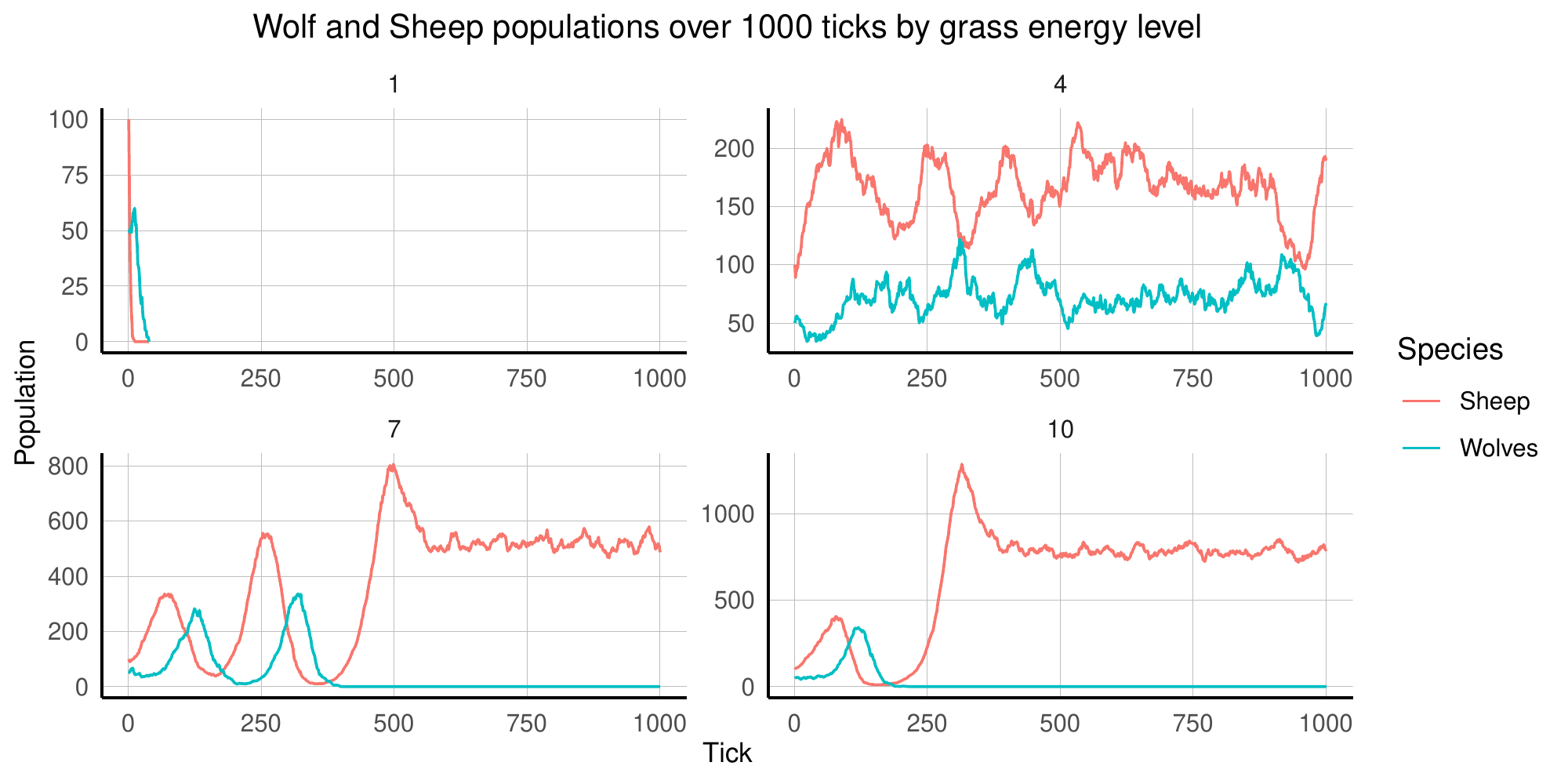}
    \caption{Four runs of Wolf-Sheep ABM with four different grass energy parameters - 1, 4, 7 and 10. Note: The y-axis is free}
\end{figure}

When the grass energy is low, both species die out extremely quickly. This is due to the sheep not gaining enough energy to move, subsequently dying out. The wolves, now with no species to predate, also die out. At a grass energy of four, an equilibrium appears to be reached. The sheep can gain enough energy to continually move, which sustains the populations of the wolves predating them. If the grass energy is changed to 7, the sheep population varies far more. The population increases sharply due to the high energy levels of the sheep (who then reproduce more frequently), which results in a high level of competition for the available grass. Due to this competition, the population drops off to an extremely low level at $\sim$340 ticks. This results in the wolf population dropping to zero from the lack of prey. With no predators remaining, the sheep population increases to a high level and remains at $\sim$500 for the remainder of the simulation. In the last run the grass energy is set to 10. This has a similar effect as the previous parameter, although the sheep population reaches a higher level before remaining constant. \\

While this \textit{`brute-force'} approach is often used to analyse complex systems for the purposes of parameter inference, it is computationally inefficient. When looping through samples of each parameter, much of the simulation time is spent producing data far from the desired outcome. For a set of parameters $\theta$ of dimension $P$, each sampled at N equal intervals $a \times \frac{max(\theta)}{N}, a = \{1,...N\}$, the required number of simulations quickly becomes prohibitively large. Consider the case of the wolf-sheep predation model with 7 parameters. If we would like to sample each parameter at 100 even intervals, the number of simulations to cover all possible combinations becomes:
\begin{align}
\text{Num Simulations} = N^{P} = 1 \times 10^{14}
\end{align}
Clearly this number of simulations is unreasonable. This can be shortened by reducing the number of sampled intervals for each parameter or disregarding some combinations of variables, but the issue remains of a large amount of computation time being wasted sampling parameters which do not produce simulated data close to the desired outcome of the ABM.

\subsection{Equation Based Modelling (EBM)}

While ABM is a 'bottom up' approach focusing on the aggregate behaviour of agents, Equation Based Modelling (EBM) is a top down approach. EBM is based on an interrelation between a series of partial differential equations (Parunak, Savit \& Riolo, 1998). The variability in the system is examined over time and/or space. Validation is enabled by comparing the output of the model to the real world system behaviour. In comparison, ABM allows validation of both the individual and aggregate system behaviour. EBM is suited for modelling physical processes where the behaviour of the components is not as important as that of the overall system. \\

Generally, EBM's are a set of differential equations and parameters. Extending the \textit{Lotka--Volterra} model, time can be eliminated from each equation to produce a single equation, and then solved for a constant quantity $V$:
\begin{align}
\frac{dx}{dt} = \alpha x - \beta x y \\
\frac{dy}{dt} = \delta x y - \gamma y \\
\frac{dy}{dx} = -\frac{y}{x} \frac{\delta x - \gamma}{\beta y - \alpha} \\
\frac{\beta y - \alpha}{y} dy + \frac{\delta x - \gamma}{x} dx = 0 \\
V = \delta x - \gamma \ln(x) + \beta y - \alpha \ln(y)
\end{align}
This model will give the number of two populations $x$ and $y$ given the set of parameters $\alpha$, $\beta$, $\gamma$ and $\delta$. A major limitation of this in ecological models is that neither population can reach zero - they always recover after falling to a low level. However, the more flexible structure of an ABM allows either population to go extinct.

\subsection{Comparison of ABM to EBM}

While Agent Based Modelling is well suited for many classes of complex systems, it is better suited for some contexts than others. In problems with a large number of homogeneous agents, it is often the case that more accurate results can be obtained in a shorter amount of time using an aggregate solution. Tracking the behaviour of individual agents is not always necessary. For example, if we are concerned with the amount of water in a container, it is more useful to track the water level directly and model it using a set of differential equations. Parameters in the equations such as temperature can be adjusted and the effects observed. Where there are only a few interacting agents in a system, a similar approach can be taken by deriving a model with differential equations. One such example is in the behaviour of a tennis ball falling onto different surfaces. Here, the physical behaviours of the model are understood with a rigorous mathematical structure such as kinematic equations. This can be used to developed a set of differential equations for effectively describing the behaviour of the ball given different materials and surfaces.\\

Agent Based Modelling is well suited for systems with many heterogeneous agents interacting in an environment. In many populations, individual agents possess unique properties which are not shared across the entire population. Their behaviours are guided by both the environment, the history of their actions and their current properties. ABM is well suited for a system where it is useful to track individual agents and their actions. In this manner it is much more powerful than using equation based modelling. Where the behaviour of agents relies on their history of interactions, the agents can be modelled as intelligent beings with the ability to learn, reason about situations, and deal with uncertainty. An example of this is in the behaviour of pedestrians in a transportation network. Humans are intelligent beings, with each individual's decision making process different from the rest. The reason for one individual choosing to travel a certain route is likely different to another individual, and the long term behaviours of each pedestrian differ substantially. While it would be impractical to model such a process using EBM, pedestrian modelling is well suited to an ABM approach, allowing for individual actions carried out by several heterogeneous agents.

\subsection{Agent Intelligence}
Recent advances in the field of artificial intelligence have allowed for the construction of agents in an ABM, such that they are able to learn from their past and take an action based on a complex set of variables. One such influential paradigm to enable this is BDI - Belief-Desire-Intention (Singh, Padgham \& Logan, 2016). BDI recognises ``that an agent can be identified as having: a set of beliefs about its environment and about itself; a set of desires which are computational states which it wants to maintain, and a set of intentions which are computational states which the agent is trying to achieve" (O'Hare, Jennings, 1996). More recent developments have expanded on this philosophy, going beyond simple reactive behaviour to embedding cognitive complexity (Abrams, 2013). While many ABMs are not suited to this more complex agent intelligence, the recent advances in modelling highly complex systems allows for the embedding of more advanced levels of agent intelligence to achieve a more realistic simulation.

\subsection{Mathematical Background of ABM}

Although ABM is much less mathematically oriented than EBM, it is important to establish a mathematical framework for representing the system and the agent properties. There exist a number of mathematical formalisations for Agent Based Models. A series of mathematical objects can be developed to represent a theoretical abstraction of agent based simulations. These objects capture the features of simulations and allow for the development of mathematical theory to explain the output of a simulation (Laubenbacher et al, 2007). An ABM can be described as a Markov Chain where the state of the system at time $t$ is given by the state at all nodes in the system: $X_{t} = {x_{i,t}},\; i = 1,...,N, t = 1,...,T$.
The state at a given node $i$ for time $t+1$ is given by
\begin{align}
x_{i,t+1} = f_{i}(X_{t}, \Xi_{t}, \theta)
\end{align}
That is, the state at time $t+1$ in the system is reliant only on the previous state. This is an important feature of an ABM, allowing for the model at each tick of the simulation to make agent decisions and update the environment based only on the current state. Storing every prior state and set of decisions, while updating the current state using this prior data, becomes costly in terms of computation and memory as a system increases in complexity. \\

Perhaps the most common framework for describing agent based simulations is Cellular Automata (CA). They are typically defined over a regular grid such as the two dimensional $\Z^{2}$. Each grid point $(i,j)$ represents a \textit{site} or \textit{node}. Each node has a state $x_{i,j}(t)$, where $t$ denotes a time step. The neighborhood \textit{N} for each site consists of a set of nodes that can influence the future state of the node $x_{i,j}$. Based on the current state $x_{i,j}(t)$ and the current state of all nodes in the neighborhood $N$, a function $f(i,j)$ computes the next state $x_{i,j}(t+1)$ of the site at $(i,j)$. Laubenbacher et al (2007) defines this as:
\begin{align}
x_{i,j}(t+1) = f(i,j)(\bar{x}_{i,j}(t))
\end{align}
where $\bar{x}_{i,j}$ denotes the tuples consisting of all the states $x_{i',j'}(t)$ with $(i',j') \in N$. \\

Another modelling framework is that of \textit{finite dynamical systems}. This framework represents an ABM as a time-discrete dynamical system of a finite state set. It can be seen that the representation of an ABM in this way is mathematically rich enough to allow the derivation of formal results.

\subsubsection*{Finite Dynamical Systems}
A Finite Dynamical System (FDS) is an iteration of a function over a collection of variables. We denote this collection as a nonempty, finite set $X = x_{1}, ... , x_{n}$. There is a single, local function $f_{i}$ associated with each element $x_{i}$ in the set. Each of theses local functions take input from only variables in the `neighborhood' of $x_{i}$. Laubenbacher et al (2007) defined an FDS as a sequential composition of these local functions, forming the dynamical system: 
\begin{align}
\Phi = (f_{1},...,f_{n}): X^{n} \to X^{n},
\end{align}
Here, $\Phi$ is an iteration unit which generates the dynamics of the system. From Laubenbacher et al (2007), $\Phi$ is assembled from each local function $f_{i}$, each of the variables can be updated simultaneously:
\begin{align}
\Phi(x_{1},...,x_{n}) = (f_{1}(x_{1},...,x_{n}),...,f_{n}(x_{1},...,x_{n}))
\end{align}
In the above a \textit{parallel dynamical system} is obtained. Each variable $x$ in the system is updated in parallel using a local function in its own neighborhood. Alternatively, a \textit{sequential} dynamical system can be obtained by updating the variables states by a fixed update order (Laubenbacher et al, 2007):
\begin{align}
\Phi_{\pi} = f_{\pi t} \circ f_{\pi n-1} \circ ... \circ f_{\pi 1},
\end{align}
where $\pi$ is a permutation on a set $\{1,...n\}$.
The dynamics of $\Phi$ can be represented as a directed graph on the vertex set $X^{n}$, called the \textit{phase space} of $\Phi$ (Laubenbacher et al, 2007). There is a directed edge from \textbf{v} $\in X^{n}$ to \textbf{w} $\in X^{n}$ if and only if $\Phi(\textbf{v})$ = \textbf{w} (Laubenbacher et al, 2007). \\

An Agent Based Model can be represented using this framework. Agents can be thought of as entities in a finite dynamical system, each carrying a set of configurations, preferences and spatial and/or temporal state information. \textit{Cells} are features of the environment in which the agents carry out behaviours. These cells take on one of a finite number of possible states. Agents interact with a subset of nearby agents in the environment, and update their internal state according to a function at each time step. The neighbors of a cell in the ABM environment form an \textit{adjacency relation}, obtaining a dependency graph of the agents (Laubenbacher et al, 2007). For example in a pedestrian network, the state at time $t$ for a given intersection $I$ is dependent on the states of the intersections in the neighborhood of $I$ at time $t-1$. \\

The agents in an ABM may be updated in a number of ways. These are generally either synchronous, asynchronous, or event-driven (Laubenbacher et al, 2007). The choice of the scheduling system is highly dependent on the system that is being modelled. In the pedestrian model described previously, it is clear that at each time-step (or \textit{tick}) of the simulation, all agents will update; moving through the environment and updating their internal states. This is a \textit{synchronous} system. An asynchronous system is suitable for models where the actions of each agent do not depend on the current state of the agents in its local neighborhood. The choice of such a scheduling system is an important feature in the design of an ABM. The nature of dependency and interaction must be carefully considered in order to design a realistic system.

\subsection{Link to statistical theory}

As the foundation of Agent Based Modelling lies in simulation based on observed data in addition to data output, statistics plays a major role at nearly every stage of the modelling process. Perhaps the most important step in ABM based research lies in the analysis of data from the simulation to derive parameters for understanding the data generating process. Other analysis may include descriptive, predictive, or inferential statistics. These models can include, but are not limited to; spatial, network, survival, or time series analysis. Most ABM frameworks provide a method to import and export data into more robust statistical packages such as R or Python. These packages provide advanced modelling tools which allow the researcher to extract insights from the ABM.

\subsection{Model Design}
To date, there is no standard format for the documentation of an ABM. Limited replicability is an issue in scientific research using ABMs. To maximise the reproducibility of an ABM, a format for their design should be adopted which is popular in the scientific community. By standardising the description of a model in this way, it can be more easily reproduced by another party or revisited by the original developer. One such format is ODD - Overview, Design, Details, first proposed by Grimm et al (2006) as a standard protocol for describing agent based models. It was subsequently reviewed and updated in 2010 (Grimm et al, 2010). ODD describes a set of key aspects of the model:

\begin{figure}[H]
    \centering
    \includegraphics[width=15cm]{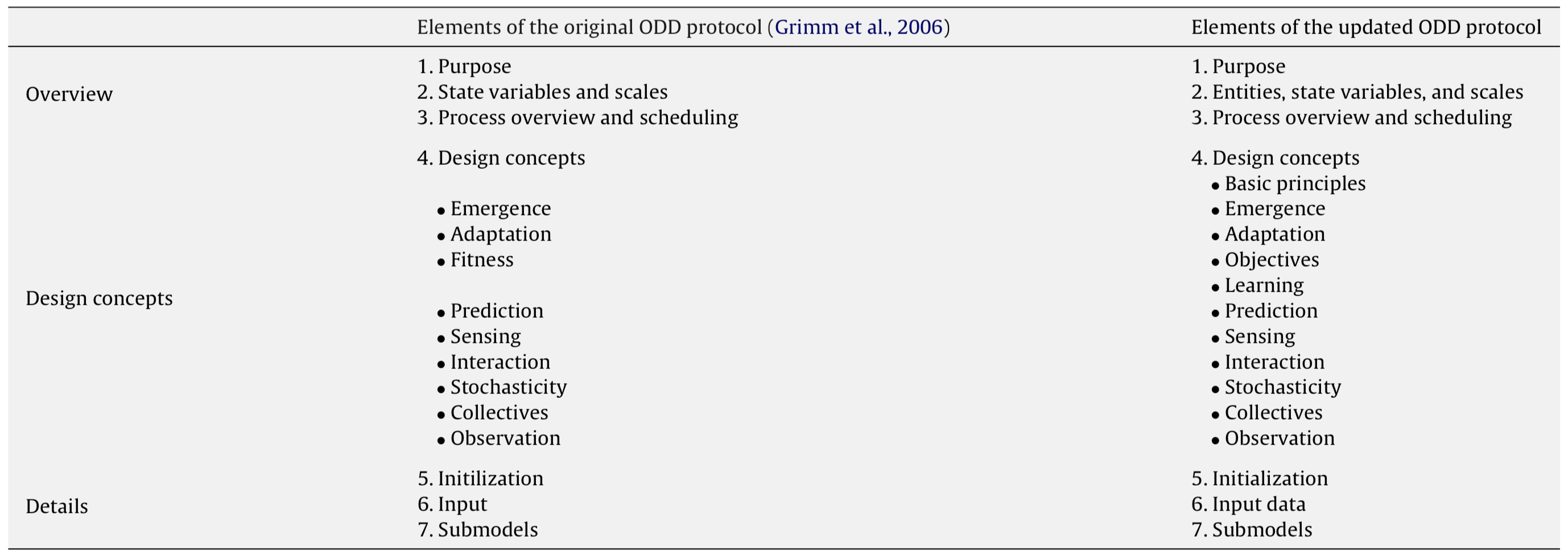}
    \caption{The ODD protocol and the changes from 2006 to the updated 2010 protocol}
    \vspace{-1.5em}
\end{figure}

\begin{enumerate}
\item{\textbf{Purpose}} \\
What is the purpose of the model? What are the desired outcomes?

\item{\textbf{Entities, State Variables and Scales}} \\
What entities exist in the model? Agents, environment, collectives, spatial units. How are the entities characterised? What properties do they hold? What are the spatial and temporal resolutions of the model?

\item{\textbf{Process overview and scheduling}} \\
What actions are assigned to each entity, and in which order? How are the state variables updated? How is time modelled - discretely, continuously, or as a continuum where both can occur?

\item{\textbf{Design Concepts}} \\
Emergence, Adaptation, Fitness, Prediction, Sensing, Interaction, Stochasticity, Collectives, Observation.

\item{\textbf{Initialisation}} \\
What is the initial state of the model world? How many entities are created and what are their default properties?

\item{\textbf{Input Data}} \\
Is external data used as input to represent the change in a process over time?
\end{enumerate}

ABMs developed in the course of this dissertation will follow the ODD protocol. For each, thorough documentation is provided to enable reproduction of the model by researchers.

\subsection{ABM Software}
Several ABM toolkits are available to enable the development of agent based model simulations. Perhaps the most popular of these is \textbf{NetLogo}. NetLogo was developed in 1999 by Uri Wilesnky. It uses a philosophy of `low threshold, no ceiling', with a goal of making the software easy to use for beginners while remaining flexible enough to develop advanced models. It contains a number of example models for fields such as network analysis, social science, and biology. All ABM models in this report are developed in NetLogo. The RNetLogo and pyNetLogo libraries in R and Python respectively allow for programmatic, headless control of a NetLogo model. Statistical models were developed in both R and Python, with the simulation element communicating with the NetLogo model for sending commands and receiving output.

\begin{figure}[H]
    \centering
    \includegraphics[width=10cm]{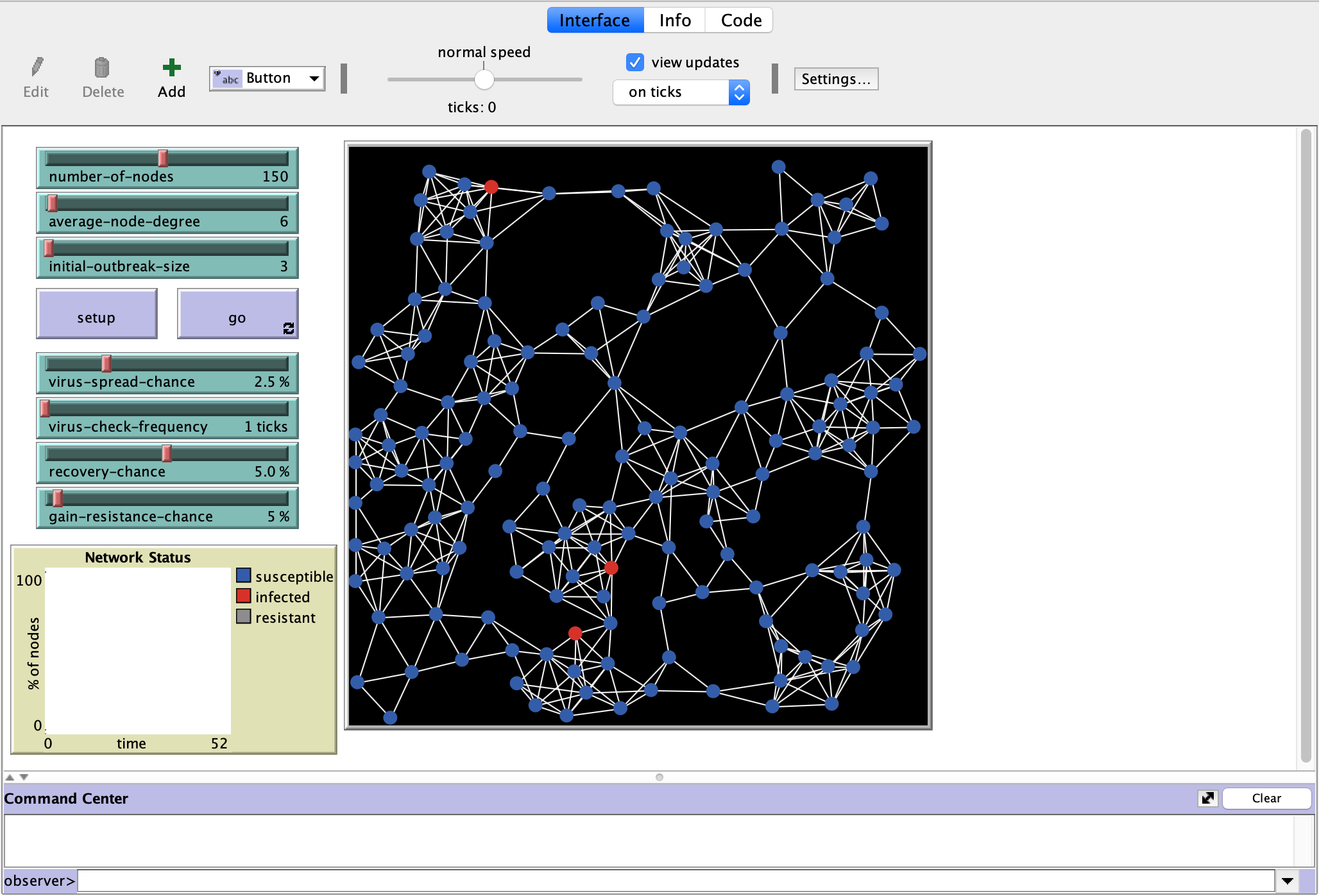}
    \caption{A NetLogo ABM of virus spread through a network}
\end{figure}

\section{Graph Theory}

ABM is commonly used to design networks in a variety of fields, including power grids - Durana et al (2014); social networks - El-Sayed et al (2012); and transport - Huynh et al (2015). Graph Theory underlies much of the construction of these networks. The basis of graph theory is the relationships between a set of entities, or \textit{nodes}. A graph is an ordered pair $G = (V,E)$ comprising a set of vertices or nodes $V$ together with a series of edges $E$, which are two-element subsets of $V$ (Prathik et al, 2016). This turns out to be a natural way of defining an ABM. For example, a model for social network analysis could be set up with users as nodes, communication between users as links, and the direction of the communication as the direction of the links. In the ABM, agents are defined as users. Their parameters may include friend count, post rate, and other metrics related to the social network in question. The environment in which the agents interact is the graph itself. Their behaviours over a series of ticks allow output such as activity levels and friend counts to be analysed on an aggregate level. \\

Consider a simple stochastic ABM represented as a graph (pictured below). In this ABM, agents are placed randomly at nodes. At each tick of the simulation each agent will move to a connected node via the grey lines. The graph is \textit{undirected} - agents can travel to any connected edge from a given node. There are seven nodes and six edges. By designing this ABM network using the logic of graph theory, we can answer route based questions, such as the number of possible ticks it will take for an agent to move from node $6$ to node $0$. Other problems we might want to solve include the shortest possible distance between two nodes, or whether it is possible to return to a node. \\

\begin{figure}[H]
    \centering
    \includegraphics[width=5cm]{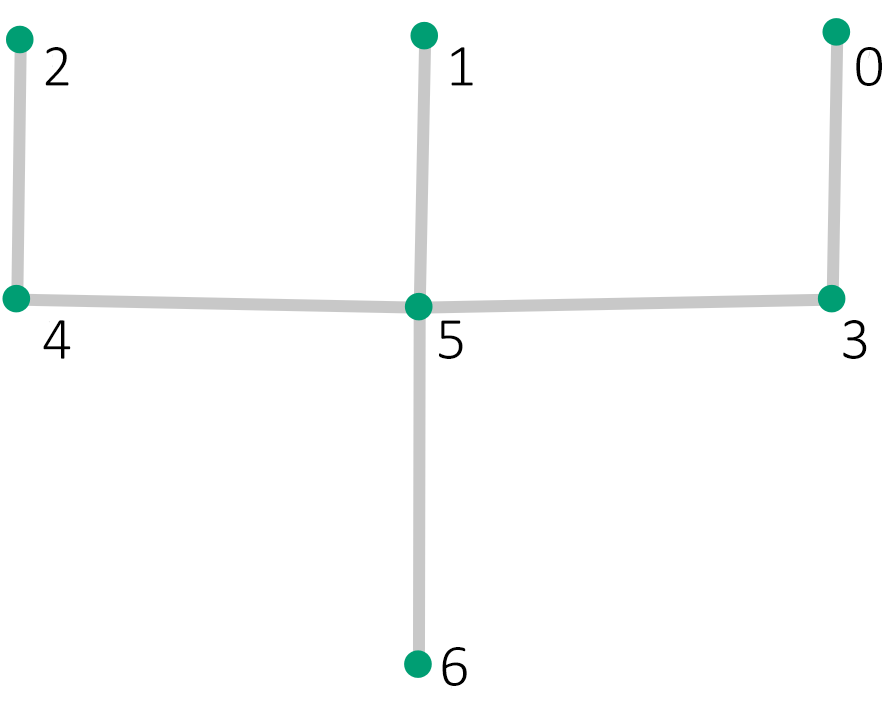}
    \caption{Undirected graph of a stochastic ABM}
\end{figure}

Constructing a network based ABM using the logic of graph theory allows for the formal analysis of the simulation. Consider an ABM consisting of a series of $N$ cities (nodes) and $R < N^{2}-N$ directed roads (links). A question that can be asked of this model is ``does there exist a tour of all $N$ cities stopping at each city exactly once using the given roads?" (Altschuler \& Williams, 2017). This is a route problem known as the \textit{Hamiltonian path problem}. This is an NP-complete problem that is solved in non-deterministic polynomial time. The model can be developed in an ABM structure to set up a simple rule based system for the traversal of a single agent between nodes. Various algorithms based on the graph theory analysis of the problem can be implemented as actions by the agent based on its current state. Setting up the model in this way allows for both abstraction from its mathematical form into one which is easier to visualise, as well as an environment to enable sandbox testing of different graph structures. This simplification is a common thread in the design of ABMs and is a large reason for their widespread use in network analysis.

\section{Approximate Bayesian Computation}

\subsection{Bayesian Inference}
Bayesian inference is a method of statistical inference grounded in Bayes Rule. It states that the probability of some event A, given that event B has occurred, is given by:
\begin{align}
P(A|B) = \frac{P(B|A)P(A)}{P(B)}
\end{align}
In the context of Bayesian parameter estimation, we aim to find the posterior distribution of the parameters given the data:
\begin{align}
p(\theta|x) = \frac{p(x|\theta) p(\theta)} {p(x)},
\end{align}
where $p(\theta|x)$ is the posterior distribution, \\
$p(x|\theta)$ is the likelihood, \\
$p(\theta)$ is the prior, and \\
$p(x)$ is the probability of the data. Often the denominator $p(x)$ involves computing an integral to normalise the posterior to a probability distribution:
\begin{align}
p(x) = \int p(x|\theta)p(\theta)d \theta
\end{align}
In some cases, such as models where a conjugate prior is available, this is straightforward to calculate numerically. If no conjugate prior is available, there are a wide range of computational methods for sampling from the posterior. One of these is Markov chain Monte Carlo (MCMC). MCMC constructs a Markov chain with the target posterior dsitibution as its equilibrium distribution. By sampling from this chain for a long period of time, a sample from the posterior is obtained. A common algorithm for this is known as Metropolis Hastings, developed by Metropolis, Rosenbluth, Rosenbluth, Teller and Teller (1953). Formally, the algorithm uses a proposal distribution $g(x^{'}|x)$ as the conditional probability of accepting a state $x^{'}$ given $x$, and an acceptance ratio $A(x^{'}|x)$ as the probability of accepting the proposed state $x^{'}$.
\begin{align}
P(x^{'}|x) = g(x^{'}|x)A(x^{'}|x), \text{where} \\
A(x^{'}|x)= \text{min}\left(1,\frac{P(x^{'})}{P(x)}\frac{g(x|x^{'})}{g(x^{'}|x)}\right)
\end{align}
MCMC can be used for parameter inference in any class of statistical model where the likelihood $g(x^{'}|x)$ can be evaluated. Consider a simple generalised linear model of the form $\pi = \frac{1}{1 + \exp[-(\beta_{0} + \beta_{1}x_{i})]}$. Here the parameters are the intercept $\beta_{0}$ and a single coefficient $\beta_{1}$. The likelihood follows a Binomial distribution of the form
\begin{align}
L(p|x) = \frac{n!}{x!(n-x)!}p^{x}(1-p)^{n-x}
\end{align}
The use of this likelihood in the calculation of the acceptance probability will construct a Markov chain that converges on the posterior distributions for $\beta_{0}$ and $\beta_{1}$.

\begin{figure}[H]
    \centering
    \includegraphics[width=15cm]{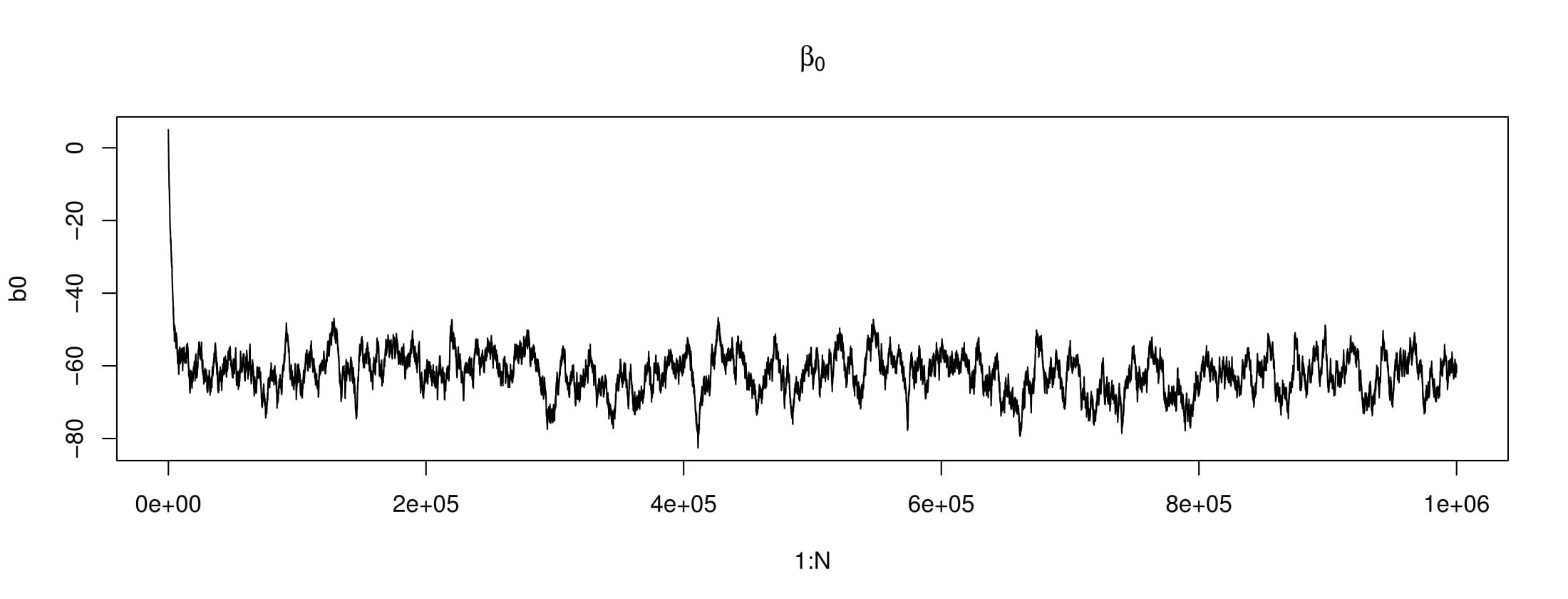}
    \caption{Convergence of the Markov Chain for $B_{0}$, following the initial burn in period}
\end{figure}

For a large class of statistical models, the likelihood is either prohibitively expensive to compute or intractable to calculate. This arises often in complex systems where stochastic processes are involved and as a result, methods such as MCMC cannot be used. An alternative method of approximating the posterior distribution is required for these problems, known as \textit{likelihood free inference}.

\subsection{Likelihood Free Inference}
The likelihood in a statistical model is a function of the parameters given the data, $f(\theta|x)$. In simple models following a known distribution this is easily calculated. However, in complex models the integral often has no closed form solution allowing it to be evaluated in a finite number of operations. An example of this is the \textit{g and k distribution}. \\

The g-and-k distribution is known as a quantile distribution, defined in terms of its inverse cumulative distribution function. It has five parameters - $\theta = \{a,b,c,g,k\}$, and its inverse cumulative distribution function is given by (Drovandi, Pettitt, 2011):

\begin{align}
Q^{gk} (z(p);\theta) = a+b\left(1+c\frac{1-\exp(-gz(p))}{1+\exp(-gz(p))}\right)\left(1+z(p)^{2}\right)^{k}z(p)
\end{align}

For a quantile distribution such as this, the likelihood cannot be calculated analytically, only numerically. In practice, simulating from the distribution to compute the likelihood is much more straightforward than the numerical calculation. We draw from the distribution by simulating values for substitution into the above function. This idea is the basis of likelihood free methods - the acceptance or rejection of samples in a simulation based model. There is ongoing development in these methods due to the advent of big data and the use of simulation models to evaluate parameters. Two such popular methods are known as \textit{Synthetic Likelihood} and \textit{Approximate Bayesian Computation}.

\subsubsection{Synthetic Likelihood}
Bayesian Synthetic Likelihood uses a multivariate normal approximation with $n$, the number of replicated model simulations, as a single tuning parameter (Price et al, 2018). In place of the likelihood of the summary statistic it uses:
\begin{align}
p_{A,n}(S_{y}|\theta) = \mathcal{N}(s_{y}; \mu_{n}(\theta), \Sigma_{n}(\theta))
\end{align}
where $s_{y}$ is a simulated summary statistic and $\mu_{n}(\theta)$ is a known data point for the parameter $\theta$. \\

An alternative form proposed by Wood (2010) uses an auxiliary likelihood based on a multivariate normal approximation. Auxiliary parameters $\mu(\theta)$ and $\Sigma(\theta)$ are used, where $\mu \in \mathbb{R}^{d}$ and $\Sigma$ is a $d \times d$ covariance matrix of a multivariate normal distribution. These parameters are unknown, but are generally simulated from the model based on $\theta$. The estimated auxiliary function is given by Price et al, (2018) as:
\begin{align}
&\mathcal{N}(s_{y}; \mu_{n}(\theta), \Sigma_{n}(\theta)),\text{where} \\
&\mu_{n}(\theta) = \frac{1}{n} \sum_{i=1}^{n}s_{i}, \\
&\Sigma_{n}(\theta) = \frac{1}{n-1}\sum_{i=1}^{n}(s_{i}-\mu_{n}(\theta))(s_{i}-\mu_{n}(\theta))^{T}
\end{align}
Bayesian synthetic likelihood arises when this auxiliary likelihood is combined with a prior distribution on the parameter (Price et al, 2018). Following Drovandi, Pettitt, and Lee (2015), BSL samples from the following target:
\begin{align}
p_{A,n}(\theta|s_{y}) &\propto p_{A,n}(s_{y}|\theta)p(\theta), \text{where} \\
p_{A,n}(s_{y}|\theta) &= \int_{S^{n}}\mathcal{N}(s_{y};\mu_{n}(\theta),\Sigma_{n}(\theta))\prod_{i=1}^{n}p(s_{i}|\theta)ds_{1:n}
\end{align}
This produces an unbiased estimate of $p_{A,n}(s_{y}|\theta)$. The ideal BSL target will be achieved as $n \rightarrow \infty$ (Price et al, 2018).

\subsubsection{Approximate Likelihood by Simulation}
In comparison, ABC non-parametrically approximates the likelihood of the summary statistic as:
\begin{align}
p_{\epsilon,n}(s_{y}|\theta) = \frac{1}{n} \sum_{i=1}^{n}K_{\epsilon}(p(s_{y},s_{i}))
\end{align}
$p(s_{y},s_{i})$ measures the distance between the observed and simulated data with some distance metric. Often this is chosen as the Euclidean distance, $\sum_{i=1}^{n}(s_{y} - s_{i})^{2}$. As with BSL, the ABC framework will converge on the posterior distribution as $n \rightarrow \infty$.

\subsection{ABC Overview}
ABC aims to approximate a posterior distribution without the existence of the likelihood function $P(y|\theta)$. In essence this is achieved by simulating data with a set of prior parameters $\theta^{'}$. The simulated data is then compared with the observed data in some capacity - either in totality or via a set of summary statistics. If the simulated data is deemed close enough to the observed data (using a defined distance measure), the set of sampled prior parameters belongs to the approximated posterior distribution $P(\theta | d(y^{*}, y) \leq \epsilon)$. Formally this can be denoted as
\begin{align}
p(\theta|x^{*}) &= \frac{f(x^{*}|\theta)\pi(\theta)}{p(x^{*})} \\
\approx p_{\epsilon}(\theta|x^{*}) &= \frac{\int f(x|\theta)\pi (\theta)\mathbbm{1}_{\Delta(x,x^{*})\leq \epsilon}dx}{p(x^{*})},
\end{align}
where $\mathbbm{1}_{\Delta(x,x^{*})\leq \epsilon}$ denotes the set of sampled points where the distance measure meets the epsilon threshold. \\

While ABC algorithms are well grounded in mathematical theory, they make a number of assumptions and approximations. These need to be carefully assessed before considering the use of ABC for a given model.
As $\epsilon \rightarrow \infty$ and $N \rightarrow \infty$, $\theta_{sim} = \theta$. Practically, we need to set $\epsilon \neq 0$ to approximate the posterior in a reasonable amount of time. With a high dimensional  parameter space, the probability of simulating prior parameters equal to that of the posterior is extremely small. As $\epsilon$ increases, the posterior distribution shifts more towards the prior. A key component of ABC is choosing $\epsilon$ such that the posterior is calculated both accurately and within a reasonable amount of time. One such approach is to run the algorithm a large number of times and consider the top $k\%$ of the sampled parameters as the acceptance area. Another is to run the algorithm multiple times until a desired $k\%$ of $N$ is attained. \\

A second potential issue is in the use of summary statistics. In most cases the complete dataset is high in dimensionality. When comparing the observed and simulated data, most samples will be rejected and the algorithm will need to be run for an impractical amount of time. If the data is reduced to a set of summary statistics such that $dim(s) << dim(y^{*})$, this curse of dimensionality is avoided. However, a poor choice of summary statistics will result in a biased posterior that is not truly representative. Additionally, the credible intervals will be inflated due to a loss of information (``Approximate Bayesian computation," n.d.). This can be resolved by choosing a \textit{sufficient statistic}, which completely explains the original dataset using a smaller subset of statistics.

\subsubsection{Summary Statistics}
When reducing the dimensionality of a sample with a summary statistic, care must be taken to choose summary statistics that produce an unbiased approximation of the posterior distribution. R.A. Fisher (1922) defined a summary statistic as  \textit{sufficient} if ``no other statistic which can be calculated from the same sample provides any additional information as to the value of the parameter to be estimated". If the summary statistic is a sufficient statistic, the posterior distribution will be equivalent to the posterior distribution computed with the full dataset, given by:
\begin{align}
f(y^{*}|\theta) \propto f(s(y^{*})|\theta),
\end{align}
where $s(y^{*})$ is a sufficient statistic for the data $y^{*}$. \\

Let $D=\{X_{1},...,X_{n}\}$ be a random sample from a probability distribution with some parameter $\theta$. A summary statistic $S(D)$ is sufficient for a parameter $\theta$ if the conditional probability distribution of the full data $D$ given $S(D);\theta$, does not depend on the parameter $\theta$ (Aeschbacher, Beaumont \& Futschik, 2012).
\begin{align}
\pi(D=d|S(D) = s,\theta) = \pi(D=d|S(D)=s)
\end{align}

 In the absence of a sufficient statistic, which is often the case in a complex system, a tradeoff must be found between a loss of information and dimensionality reduction. The incorrect choice of summary statistics will result in a biased posterior which is not representative of the data. Methods exist to find summary statistics which contain the maximum amount of information about the parameters of interest. One such method is \textit{Partial least squares regression}. \\

 Partial least squares (PLS) is a projection technique which ``aims to produce linear combinations of covariates which have high covariance with responses and are uncorrelated with each other" (Sisson et al, 2018). In the context of ABC, the covariates are denoted as $z_{1},...,z_{k}$, and the responses as $\theta_{1},...\theta_{p}$. The \textit{i}$^{th}$ PLS component $u_{i} = \alpha_{i}^{T} z$ maximises
 \begin{align}
\sum_{j = 1}^{p}\text{Cov}(u_{i},\theta_{j})^{2},
 \end{align}
subject to Cov($u_{i},u_{j}$) = 0 for $j < 1$ and a normalisation constraint on $
\alpha_{i}$ such that $\alpha_{i}^{T}\alpha_{i}$ = 1 (Sisson et al, 2018). The resulting components can be viewed as a lower dimensional set of statistics than the original data, and can hence be used in ABC in place of the original data for more efficient inference. An advantage of projection methods such as PLS is in their interpretability. It is easy to understand how maximising the variance of dimensions in a dataset $D$ into a subset $z$ will often result in a sufficient representation of $D$. \\

Another set of techniques to reduce the dimensionality of the data are known as \textit{subset selection methods}. Joyce and Marjoram (2008) propose a step-wise selection approach, where candidate summary statistics are added and/or removed to a subset, and the effect evaluated on the ABC posterior. The rationale behind this is if the set of statistics are sufficient, adding further such statistics will not affect the likelihood of the resulting posterior; however removing statistics will. From Joyce and Marjoram (2008), the technique deems a change to the subset as significant if:

\begin{equation}
\left|\frac{\hat{\pi}_{ABC}(\theta|S^{'}(y_{obs}))}{\hat{\pi}_{ABC}(\theta|S(y_{obs}))}-1 \right|>T(\theta),
\end{equation}
where $\hat{\pi}_{ABC}$ is an estimated posterior density based on the output from rejection ABC (Sisson et al, 2018). \\

There exist many other techniques to reduce a large dataset to a smaller set of summary statistics, in addition to ongoing research into new techniques. Each of them have strengths and weaknesses, and therefore should be carefully evaluated when using them in a particular model. 

\subsection{Rejection ABC}
The simplest ABC algorithm, rejection sampling, is a straightforward procedure of simulating data with some given set of parameters and evaluating the data against a set of observed data. The algorithm proceeds as follows:

\begin{algorithm}[H]\caption{ABC Rejection}\label{abc-rej}
\begin{algorithmic}[1]
\State Compute a summary statistic $s(y)$ from the observed data
\State Define a distance measure $d(s(y), s(\hat{y}))$
\State Define an distance threshold $\epsilon$
\For  {$i=1$ to $N$} 
\State Sample a set of parameters from the prior distribution $\theta^{*} \sim \pi(\theta^{*})$
\State Simulate a dataset given the sampled parameters: $x^{*} \sim \pi(x^{*}|\theta^{*})$
\State Compute a summary statistic $s(\hat{y})$ for the simulated data
\State If $d(s(y), s(\hat{y})) < \epsilon$, keep the parameters $\theta^{*}$ as a sample from $\pi(\theta | x)$, otherwise reject
\EndFor
\State Approximate the posterior distribution $\theta$ from the distribution of accepted parameter values $\hat{\theta}$
\end{algorithmic}
\end{algorithm}

Direct evaluation of the likelihood is not carried out in the above algorithm. For almost any model where data is observed, data can be simulated, and there exists a set of parameter values to be inferred, rejection ABC can be used as a straightforward approach to approximate the posterior distribution. This is achieved by noting that the acceptance probability of a given parameter is proportional to the probability of generating the observed data $y_{obs}$, under the model $p(y|\theta)$ for a fixed parameter vector $\theta$. For a fixed $\theta$, if we generate a dataset from the model $y\sim p(y|\theta)$, then the probability of generating our observed dataset exactly, so that $y=y_{obs}$, is equal to $p(y_{obs}|\theta)$ (Sisson, 2018).

\subsubsection{Regression Adjustment in Rejection ABC}

Regression adjustment is a technique commonly used in Bayesian methods where sampling techniques are proposed to account for discrepancies between simulated and observed summary statistics (Blum, 2017). In the context of ABC, the set of posterior parameters following the simulation are adjusted using weights proportional to the distances between the simulated and summary statistics. This approach gives weighting to parameters which give simulated data closer to the observed data. The parameter values are adjusted by fitting a regression model of the relationship between the posterior parameters and the simulated data (Francious, 2010):
\begin{align}
\theta_{s} = \alpha + \beta(y-y_{s}) + \mathcal{E}_{s}, s = 1,...,N_{\epsilon}
\end{align}

The linear model is fit as above, and the parameters are corrected as follows (Francious, 2010): 
\begin{align}
\theta^{*}_{s} = \hat{\alpha} + \mathcal{E}_{s} = \theta_{s} = \hat{\beta}(y-y_{s})
\end{align}

This method can result in significantly improved posterior estimates for rejection ABC. When $y$ is high in dimensionality, non-linear regression adjustment is often favoured, through a number of techniques such as GAM's, ridge regression and feed-forward neural networks (Francious, 2010).
\subsection{ABC-SMC}

A limitation of the rejection algorithm is a low acceptance rate when the prior is very different to the posterior. Beaumont et al (2009) proposed a particle filtering based method known as ABC-SMC. This algorithm alleviates the issue of low acceptance rates by avoiding low rejection rates of the sampling region and gradually evolving towards the target posterior through population filters. In high dimensional data from a complex system this results in a large gain in efficiency. The algorithm proceeds as follows: \\

\begin{algorithm}[H]\caption{ABC-SMC (Beaumont et al, 2009)}\label{abc-smc}
\begin{algorithmic}[1]
\State Initialize threshold schedule $\epsilon_{1} > ... > \epsilon_{T}$
\State Set $t = 1$
\For  {$i=1$ to $N$} 
\State Simulate $\theta_{i}^{(1)} \sim p(\theta)$ and $x \sim p(x|\theta_{i}^{(1)})$ until $p(x,x_{obs}) < \epsilon_{1}$
\State Set $w_{i} = 1/N$
\EndFor
\For  {$t=2$ to $T$} 
\For {$i = 1,...,N$}
\State Repeat:
\Indent
\State Pick $\theta_{i}^{*}$ from the $\theta_{j}^{(t-1)}$'s with probabilities $v_{j}^{(t-1)}$,
\State Draw $\theta_{i}^{(t)} \sim K_{\theta,t}(\theta_{i}^{(t)}|\theta_{i}^{*}$ and $x \sim p(x|\theta_{i}^{(t)});$
\EndIndent
\State until $p(x,x_{obs}) < \epsilon_{t}$
\State Compute new weights as
\State \begin{align}
w_{i}^{(t)} \propto \frac{p(\theta_{i}^{(t)})}{\sum_{j}v_{j}^{(t-1)}K_{\theta,t}(\theta_{i}^{(t)}|\theta_{j}^{(t-1)})}
\end{align}
\State Normalise $w_{i}^{(t)}$ over $i = 1,...,N$
\EndFor
\EndFor

\end{algorithmic}
\end{algorithm}

$K_{\theta,t}$ in step 14 of the algorithm is chosen as a conditional density that serves as a transition kernel to “move” sampled parameters and then appropriately weight accepted values. In contexts of real-valued parameters, for example, $K_{\theta,t}(\theta|\theta^{*})$ might be taken as a multivariate normal or $t$ density centred at or near $\theta^{*}$, and whose scales may decrease as $t$ increases. (Bonassi \& West, 2015). Kernel choice is discussed further in Section 5.5.1.\\

The algorithm constructs $T$ intermediary distributions with increasingly small tolerance schedules. The tolerances $\epsilon$ are chosen such that gradual evolution towards the target posterior is achieved. The algorithm continues until $N$ particles have been accepted in population $T$. The final set of particles is an approximation of the posterior distribution. The inclusion of a weight calculation for each particle in a population enables the sampling from population $t-1$ with probabilities equal to the normalised weights. These weights are calculated such that particles further from the population mean are sampled and refined. \\

The ABC-SMC algorithm addresses the main drawback of rejection ABC in the inefficiency of sampling parameters with a high distance from the posterior, and hence rejecting a large number of samples. The repeated sampling from posterior approximations results in a distribution closer to the posterior distribution (Beaumont, 2010). One drawback exists in the decreasing epsilon tolerance. If the quantile of the distances from which to reduce epsilon is appropriately chosen, the number of accepted samples in each population will not decrease significantly as $\epsilon$ decreases. However, setting this quantile too high will reduce $\epsilon$ such that a larger proportion of samples are rejected, eliminating a key advantage of the SMC approach. This can be addressed by either calculating $\epsilon$ at each population using an adaptive method (described in Section 5.5.3), or including conditional density estimation for the final population sample. \\

An extension to the original SMC algorithm was proposed by Bonassi \& West (2015). A joint kernel $K_{t}(x,\theta|x^{*},\theta^{*})$ is used on the joint distribution of accepted values $(x,\theta)$ to raise the importance of proposals where the simulated data $x_{i}$ is close to $x_{obs}$. This addition is known as ABC-SMC with Adaptive Weights. The algorithm proceeds as follows:

\begin{algorithm}[H]\caption{ABC-SMC with Adaptive Weights (Bonassi \& West, 2015)}\label{abc-smc}
\begin{algorithmic}[1]
\State Initialize threshold schedule $\epsilon_{1} > ... > \epsilon_{T}$
\State Set $t = 1$
\For  {$i=1$ to $N$} 
\State Simulate $\theta_{i}^{(1)} \sim p(\theta)$ and $x \sim p(x|\theta_{i}^{(1)})$ until $p(x,x_{obs}) < \epsilon_{1}$
\State Set $w_{i} = 1/N$
\EndFor
\For  {$t=2$ to $T$} 
\State Compute data based weight $v_{i}^{(t-1)} \propto w_{i}^{(t-1)}K_{x,t}(x_{obs}|x_{i}^{(t-1)})$
\State Normalise weights $v_{i}^{(t-1)}$ over $i = 1,...,N$
\For  {$i=1$ to $N$} 
\State Repeat until $p(x,x_{obs}) < \epsilon_{t}$:
\Indent
\State Pick $\theta_{i}^{*}$ from the $\theta_{j}^{(t-1)}$'s with probabilities $v_{j}^{(t-1)}$,
\State Draw $\theta_{i}^{(t)} \sim K_{\theta,t}(\theta_{i}^{(t)}|\theta_{i}^{*}$ and $x \sim p(x|\theta_{i}^{(t)});$
\EndIndent
\State Compute new weights as
\State \begin{align}
w_{i}^{(t)} \propto \frac{p(\theta_{i}^{(t)})}{\sum_{j}v_{j}^{(t-1)}K_{\theta,t}(\theta_{i}^{(t)}|\theta_{j}^{(t-1)})}
\end{align}
\EndFor
\State Normalise $w_{i}^{(t)}$ over $i = 1,...,N$
\EndFor

\end{algorithmic}
\end{algorithm}

The added step of updating the data based weights adds computation time, although this is usually negligible in comparison to the simulation time (Bonassi \& West, 2015). The addition of adaptive weights increases the proportion of accepted samples in comparison to Algorithm 2, and hence decreases the overall number of simulations. As the simulations tend to dominate computation time, ABC-SMC with Adaptive Weights tends to be more computationally efficient than ABC-SMC.

\subsubsection{Choice of Pertubation Kernel in ABC-SMC}
From each time step $t = 2,...T$, the weights at population $t$ are calculated using a pertubation kernel:
\begin{align}
w_{i}^{(t)} \propto \frac{p(\theta_{i}^{(t)})}{\sum_{j}v_{j}^{(t-1)}K_{\theta,t}(\theta_{i}^{(t)}|\theta_{j}^{(t-1)})},
\end{align}
where $p(\theta_{i}^{(t)})$ is the probability of the sampled parameter given the prior, and \\
$K_{\theta,t}(\theta_{i}^{(t)}|\theta_{j}^{(t-1)})$ is the probability of the sampled parameter given the previous population. \\

A balance needs to be obtained in this kernel such that the parameter space is sufficiently explored, but not so extensively to cause a low acceptance rate (Filippi, 2016). The properties of an optimal kernel are derived from sequential importance sampling theory. Similarity is assessed between two joint distributions $(\theta^{(t-1)},\theta^{(t)})$ where $\theta^{(t-1)} \sim p_{\epsilon_{t-1}}$. $\theta^{(t)}$ is constructed by pertubing $\theta^{(t-1)}$ and accepting it according to some threshold $\epsilon_{t}$, which is reduced at each time step. Resemblance between the two distributions is the first criteria of an optimal kernel. An optimisation problem is solved to balance sufficiently exploring the parameter space while maintaining a high acceptance rate (Filippi, 2016). Filippi (2016) gives this as the solution to:
\begin{align}
& 1.\quad q_{\epsilon_{t-1},\epsilon_{t}}(\theta^{(t-1)},\theta^{(t)}|x) \\
& = \frac{p_{\epsilon_{t-1}}(\theta^{(t-1)}|x)K_{t}(\theta^{(t)}|\theta^{(t-1)})\int f(x|\theta^{(t)})\mathbbm{1}(\Delta(x^{*},x)\leq \epsilon_{t})dx}{\alpha(K_{t},\epsilon_{t-1},\epsilon_{t},x)} \\
\text{and} \\
& 2. \quad q^{*}_{\epsilon_{t-1},\epsilon_{t}}(\theta ^{(t-1)},\theta^{(t)}|x) = p_{\epsilon_{t-1}}(\theta^{(t-1)}|x)p_{\epsilon_{t}}(\theta^{(t)}|x)
\end{align}
This resemblance is in terms of the Kullback - Leibler divergence (Filippi et al, 2013).
\begin{align}
& KL(q_{\epsilon_{t-1},\epsilon_{t}};q^{*}_{\epsilon_{t-1},\epsilon_{t}}) = -Q(K_{t},\epsilon_{t-1},\epsilon_{t},x) + \log \alpha (K_{t},\epsilon_{t-1},\epsilon_{t},x) + C(\epsilon_{t-1},\epsilon_{t},x) \\
\text{where} \\
& Q(K_{t},\epsilon_{t-1},\epsilon_{t},x) = \iint p_{\epsilon_{t-1}}(\theta^{(t-1)}|x) p_{\epsilon_{t}}(\theta^{(t)}|x) \log K_{t}(\theta^{(t)}|\theta^{(t-1)})d\theta^{(t-1)}d\theta^{(t)} 
\end{align}

There are multiple choices of a kernel to achieve this balance. As with the selection of summary statistics, each should be carefully considered according to the model to be developed. When using ABC to validate and infer parameters of a complex system, it is vital to optimise the algorithm such that the kernel is not sampling from areas of the parameter space which give rise to output far from that of the observed data. On the other hand, a local pertubation kernel will not move the particles sufficiently and therefore may fail to find the true posterior. \\

The joint proposal distribution in the ABC-SMC algorithm corresponds to sampling a particle from the previous population $t-1$ and pertubing it to obtain a new particle (Filippi et al, 2013). The process typically proceeds as follows:
\begin{enumerate}
\item Sample a particle $\theta_{i}^{(t-1)}$ from the previous population $t-1$ using probabilities $w^{(t-1)}$
\item Pertube this particle to obtain $\theta_{i}^{(t)}$
\item Calculate the weight for the particle as the sum of the previous population weights multiplied by a kernel, comparing the newly pertubed particle to each particle in the previous population
\end{enumerate}

\subsubsection{Component wise pertubation kernel}

From Filippi et al (2013), for each element in the parameter vector $\theta = \{\theta_{1},...,\theta_{n}\}$, each individual component $1 \leq j \leq d$ of the vector is pertubed independently using a Gaussian distributed parameterised by mean $\theta_{j}$ and variance $\sigma^{2}_{j}$. Filippi et al (2013) shows this with the form:
\begin{align}
K_{t}(\theta^{(t)}|\theta^{(t-1)}) = \prod_{j=1}^{d} \frac{1}{\sqrt{2 \pi \sigma_{j}^{(t)}}}\exp{\{ -\frac{(\theta_{j}^{(t)}-\theta_{j}^{(t-1)})^{2}}{2 \sigma_{j}^{(t)2}} \}}
\end{align}

\subsubsection{Multivariate normal pertubation kernel}

It is often the case in models with a large number of parameters that correlation exists between some of the parameters. When using a component-wise pertubation kernel, this can lead to an inadequate reflection of the true posterior (Filippi et al, 2013). We can take into account this correlation by using a multivariate Gaussian distribution, which constructs a covariance matrix for the current population $\Sigma^{(t)}$. Again, using the Kullback-Leibler divergence allows for the calculation of an optimal covariance matrix. From Filippi et al (2013), this yields:
\begin{align}
&\Sigma^{(t)} \approx \sum_{i=1}^{N} \sum_{k=1}^{N_{0}}\omega^{(i,t-1)}\tilde{\omega}^{(k)}(\tilde{\theta}^{(k)}-\theta^{(i,t-1)})(\tilde{\theta}^{(k)}-\theta^{(i,t-1)})^{T}, \text{where} \\
&\{\tilde{\theta}^{(k)}\}_{1 \leq k \leq N_{0}} = \{\theta^{(i,t-1)} s.t \Delta (x^{*},x^{(i,t-1)}) \leq \epsilon_{t}, 1 \leq i \leq N \}
\end{align}
In practice this kernel is a popular choice in sequential sampling models, although many other options exist. Consider a set of parameters which are correlated but in a non-linear way. In this case we may wish to construct a covariance matrix for each particle to take into account the non-linear structure. There are multiple ways to do this, however for the purposes of simplicity we will restrict ourselves to using a Gaussian pertubation kernel without taking into account local correlation. For more information about local pertubation kernels, see Filippi et al (2013). \\

Consider a model where the parameter kernel $w_{i}$ and data based kernel $v_{i}$ (Step 8 of Algorithm 3) are both chosen as a multivariate Gaussian kernel. The data based and parameters weights respectively are updated as:

\begingroup
\large
\begin{align}
&f_{\mathbf{x}}(x_{1},...,x_{k}) = \frac{\exp(-\frac{1}{2}(\mathbf{x-\boldsymbol{\mu}})^{T}\boldsymbol{\Sigma}^{-1}(\mathbf{x-\boldsymbol{\mu}}))}{\sqrt{(2 \pi)^{k}|\boldsymbol{\Sigma}}|},
\end{align}
\endgroup
where $\mathbf{x}$ is a vector of proposed values, $\boldsymbol{\mu}$ is a vector of means of the previous population, and $\boldsymbol{\Sigma}$ is a positive-definite, symmetric covariance matrix.\\

As mentioned previously, this kernel will both explore the parameter space while maintaining a high acceptance rate. Each particle is sampled from the previous population and pertubed using a normal distribution. In the case that the chosen particle is resampled such that the simulated data is closer to the observed data, it will be accepted to the current population. If the observed data is further away and the parameter rejected, the particle from the previous population is `thrown out' and a new particle is resampled. The result is both a reduction in variance and convergence on the posterior mean, with the decreasing epsilon threshold acting to move the particles closer to the posterior. \\

This is demonstrated using a simple example of inferring the mean of a Gaussian distribution. The choice of pertubation kernel here involves finding the probability of the proposed $\mu$ given the prior mean and standard deviation, and dividing this by the probability of the proposed $\mu$ given the mean and standard deviation of the previous population, multiplied by the weights of the previous population. The prior mean is defined as $\theta \sim N(0,10)$. The weights are updated as:
\begin{align}
w_{i}^{(t)} &\propto \frac{p(\theta_{i}^{(t)})}{\sum_{j}v_{j}^{(t-1)}K_{\theta,t}(\theta_{i}^{(t)}|\theta_{j}^{(t-1)})}, \text{where} \\
p(\theta_{i}^{(t)}) &= f(\mu_{i}^{'(t)}|\mu_{\theta},\sigma_{\theta}), \text{and} \\
K_{\theta,t}(\theta_{i}^{(t)}|\theta_{j}^{(t-1)}) &= f(\mu_{i}^{'(t)}|\mu^{t-1},\sigma^{t-1})
\end{align}

$p(\theta_{i}^{(t)})$ is the probability of the proposed theta given the prior mean and variance, and \\
$K_{\theta,t}(\theta_{i}^{(t)}|\theta_{j}^{(t-1)})$ is a kernel function which sums the probabilities of the proposed $\theta$ given each particle in the previous population and the variance of the previous population. The function $f$ in the these kernels is the probability density function of a normal distribution. \\

\begin{table}[H]
\centering
\begin{tabular}{cccccc}
  \hline
Running Time & Known $\mu$ & Initial Epsilon & Distance Measure & Population Size \\
  \hline
201 seconds & 0.086 & 1 & $\frac{1}{N} \times \text{abs}(y^{*}-y)$ & 2500\\ 
   \hline
\end{tabular}
\end{table}

By the seventh population the algorithm has converged on the true parameter. Each subsequent population reduces the variance.

\begin{table}[H]
\centering
\begin{tabular}{rrrrrr}
  \hline
 & Iteration & $\mu$ & Epsilon & Mean Distance & Mean Loops Per N \\ 
  \hline
1 & 1 & -0.022 & 1.000 & 0.506 & 12.899 \\ 
  2 & 7 & -0.033 & 0.151 & 0.074 & 2.172 \\ 
  3 & 14 & -0.036 & 0.019 & 0.009 & 8.650 \\ 
  4 & 20 & -0.034 & 0.003 & 0.002 & 48.249 \\ 
   \hline
\end{tabular}
\end{table}

\begin{figure}[H]
    \centering
    \includegraphics[width=15cm]{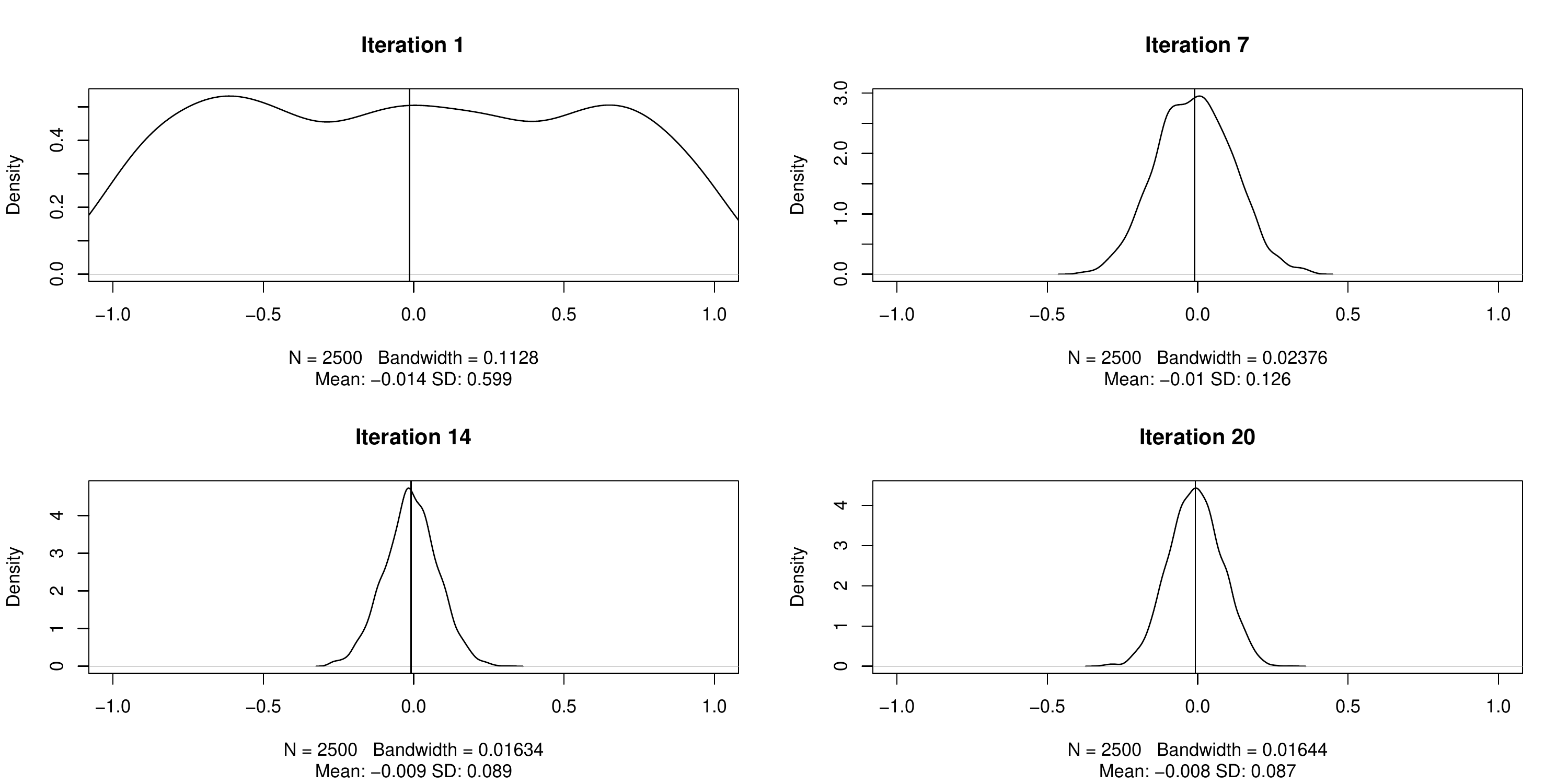}
    \caption{Progression in the posterior mean with a known $\mu$ = -0.035.}
\end{figure}

\subsubsection{Choice of Epsilon Threshold}

The approach to reducing epsilon at each population is another detail of the ABC-SMC algorithm which impacts on efficiency. Often, $\epsilon$ at each population is adjusted using a quantile based approach (Beaumont et al, 2009). For each corresponding distance vector at population $t-1$, adjust $\epsilon$ by some chosen quantile $\alpha$.
\begin{align}
\alpha\{\Delta (x^{(i,t-1)},x^{*})\}_{1 \leq i \leq N}
\end{align}
Both the choice of initial epsilon and alpha will determine the balance between computational and statistical efficiency. As mentioned previously, $\epsilon = 0$ will produce an exact posterior, but in a complex system with multiple dimensions will run for an impractical length of time. \\

One drawback to the quantile based method is that for a large alpha, the algorithm may fail to converge on the posterior. A remedy of this is to use a threshold-acceptance rate curve (Filippi, 2016). This is achieved with the following algorithm:

\begin{algorithm}[H]\caption{Adaptive method for epsilon threshold choice (S. Filippi, 2016)}\label{adaptive-epsilon}
\begin{algorithmic}[1]
\State For each population $t$,
\Indent
\State Generate a population of pertubed particles
\State Fit a Gaussian mixture model to the pertubed population
\State Estimate \begin{align}
p_{t}(x) = \int q_{t}(\theta)f(x|\theta)d\theta
\end{align}
using the unscented transform independently for each Gaussian mixture
\State Estimate \begin{align}
\alpha_{t}(\epsilon) = \int p_{t}(x)\mathbbm{1}(\Delta(x,x^{*})\leq \epsilon)dx
\end{align}
\EndIndent
\end{algorithmic}
\end{algorithm}

\subsection{Parallelising ABC}

There are three methods of controlling the balance between computational speed and posterior accuracy in the ABC-SMC algorithm - the choice of epsilon in each population, the size of the populations, and the number of populations. If the model complexity is small or the computation time is not a concern, the epsilon array can be set to small values, with a large number of populations and a large number of particles in each population. For a complex system with multiple parameters and high dimensional data, the computation time of the algorithm increases exponentially. The two principal reasons for this are 1). The time to run the simulation with the prior parameters and 2). Looping until the epsilon threshold is reached for each sample of the current population. Adjusting the two settings mentioned above will help to reduce the computation time.  \\

An additional measure is that of parallelisation. In the advent of big data there now exist several methods for conducting complex analysis on cluster computing systems. A cluster system comprises of two or more computers working together to perform tasks (ESDS, 2014). They are particularly suited to complex computational tasks which can be divided into many smaller tasks for each computer (node). For the analysis of large datasets it is important to design statistical algorithms which are well suited to this infrastructure. In some cases this is straightforward. In the rejection ABC algorithm, the number of desired loops can be divided between each node and the results merged at the end of the computation. With more complex ABC algorithms with dependencies amongst steps this is less clear. In each population of ABC-SMC there is a dependence on the distance, weight and parameter values from the previous population. Simulations within the current population $t$ can be executed in parallel, and the results sent back to the master node for evaluation until the population threshold $N$ is reached. As the simulation time dominates when using ABC to infer for an ABM, this method of parallelisation can result in a significant decrease in computation time.

\subsection{ABC Inference of a Gaussian Distribution}

Two ABC models were developed to infer the mean and standard deviation of a normal distribution - Rejection ABC and ABC-SMC with Adaptive Weights. The true mean was taken to be approximately 4. A Normal prior $\theta^{1}$ was set for the mean and a Gamma prior $\theta^{2}$ for the standard deviation.

\begin{table}[H]
\centering
\begin{tabular}{@{\extracolsep{4pt}}llccccccc}
\toprule   
 True Mean & True SD & Mean Prior $\theta^{1}$ & SD Prior $\theta^{2}$ \\
\midrule
4 & 1.5 & $\sim N(2,3)$ & $\sim Gamma(1,3)$ \\ 
\bottomrule
\end{tabular}
\caption{Normal Mean and SD Inference Model} 
\end{table}

\begin{table}[H]
\centering
\begin{tabular}{@{\extracolsep{4pt}}llccccccc}
\toprule   
 Iterations & $\epsilon$ & Acc Rate & Running Time & $\mu(\theta^{1})$ & $\sigma^{2}(\theta^{1})$ & $\mu(\theta^{2})$ & $\sigma^{2}(\theta^{2})$ \\
\midrule
30000 & 0.5 & 3.5\% & 25s & 3.973 & 0.124 & 1.46 & 0.121 \\ 
\bottomrule
\end{tabular}
\caption{Rejection ABC model} 
\end{table}

\begin{figure}[H]
    \centering
    \includegraphics[width=15cm]{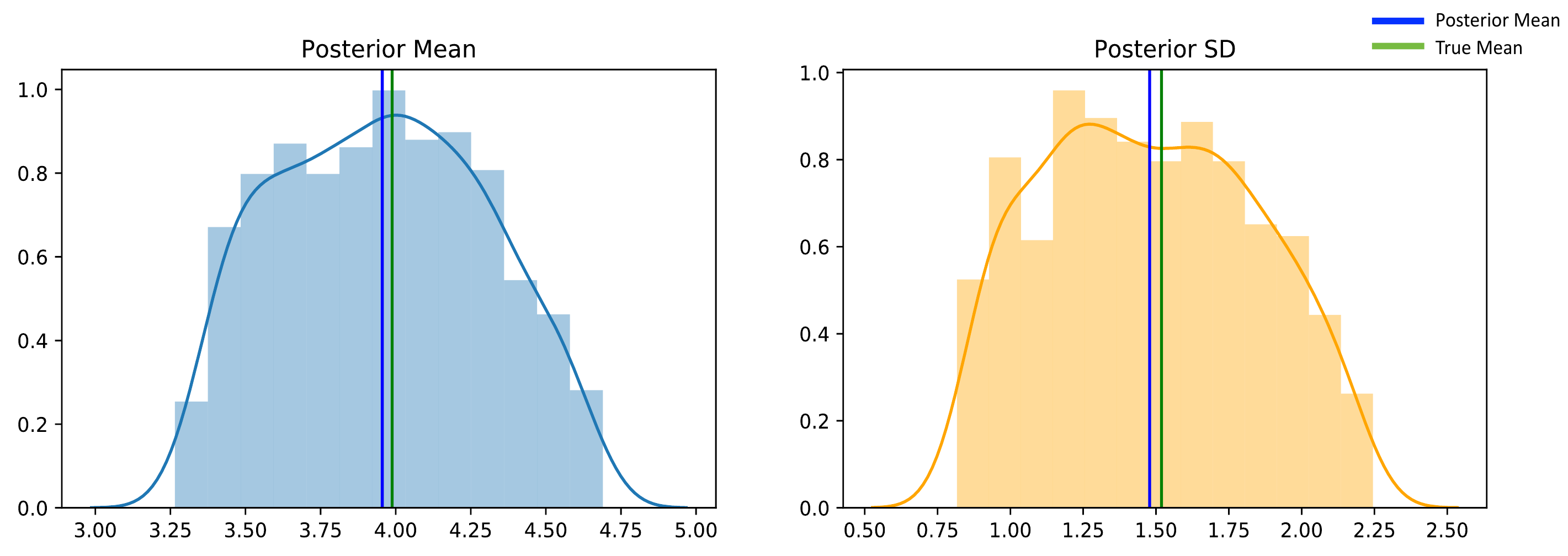}
    \caption{Posterior Distributions of the Rejection ABC Model}
\end{figure}

Rejection ABC with a epsilon of $\epsilon = 0.5$ and $N = 30,000$ results in a high rejection rate of 96.5\%. This can be adjusted through a more informative prior, although in many cases this is not available. Another adjustment is in epsilon. If this is increased, more samples will be accepted with the tradeoff of a less accurate posterior. The acceptance rate of 3.5\% appears to give an accurate posterior. 

\begin{table}[H]
\centering
\begin{tabular}{@{\extracolsep{4pt}}cccc}
\toprule   
 Populations & Iterations & Epsilon & Running Time \\
\midrule
5 & [16139, 17518, 18857, 20232] & [1, 0.74, 0.56, 0.44, 0.35] & 117s \\ 
\bottomrule
\end{tabular}
\caption{ABC-SMC model} 
\end{table}

\begin{table}[H]
\centering
\begin{tabular}{@{\extracolsep{4pt}}llccccccc}
\toprule   
$\mu(\theta^{1})$ & $\sigma^{2}(\theta^{1})$ & $\mu(\theta^{2})$ & $\sigma^{2}(\theta^{2})$ \\
\midrule
 4.006 & 0.07 & 1.50 & 0.099 \\ 
\bottomrule
\end{tabular}
\caption{ABC-SMC posterior estimates} 
\end{table}

\begin{figure}[H]
    \centering
    \includegraphics[width=15cm]{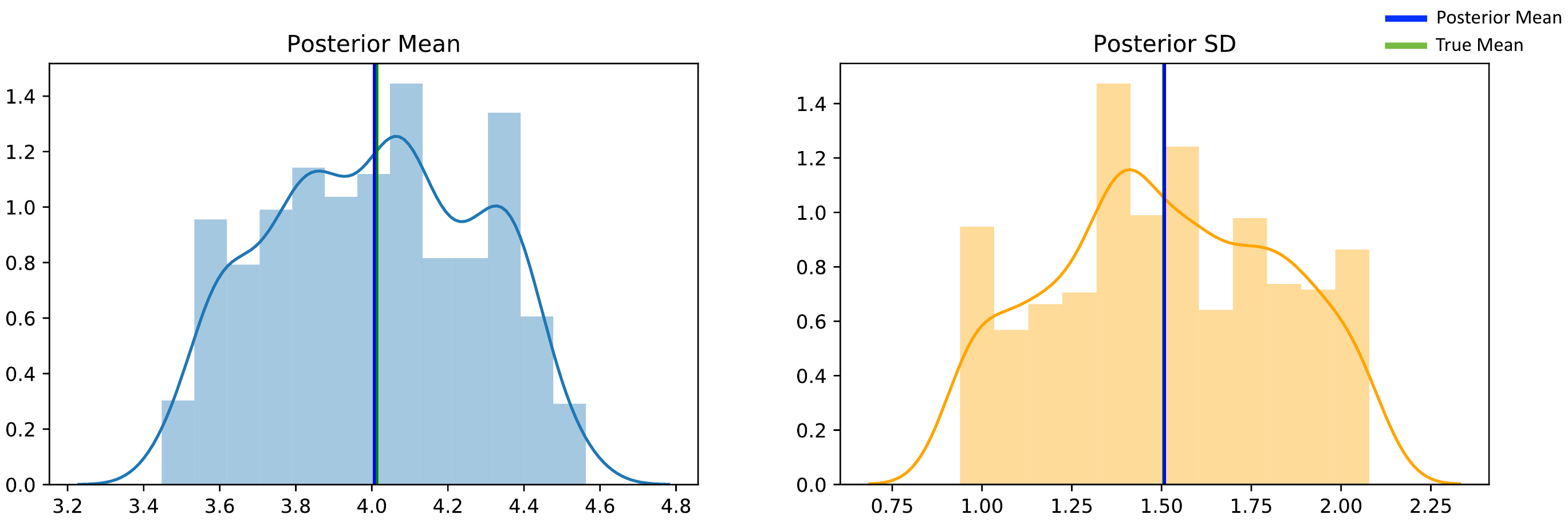}
    \caption{Posterior Distributions of the ABC-SMC Model}
\end{figure}

The ABC-SMC model has a longer running time, but reduces the variance of both posterior estimates $\mu$ and $\sigma$. The acceptance rate at the final population is 50\%, which is significantly higher than the rejection model. The increase in computation time is due to the multiple populations and increased time spent computing weights. For a complex model, the reduced time spent simulating data will outweigh this. The posterior mean estimates are slightly closer to the true mean.

\subsection{Validation of an ABM using ABC}
A challenge in designing an ABM to represent a complex system is representing the system in an accurate manner. Where there is observed data for a system, an ABM can be said to be validated against such a system if its output (simulated data) is close to the system's observed data. ABC is a suitable approach for this validation. If an ABM is designed with control from a set of parameters, ABC can be used to run the simulation with a set of prior parameters. If the ABC is modelled correctly, it will converge on a set of parameters which will produce simulated data close to the observed data from the system that the ABM is modelled from. In comparison to the `brute force' method described in Section 3.1, significantly less time is spent simulating data using parameters far from the true parameters of the system. Further models will focus on the use of ABC algorithms to infer parameters for an Agent Based Model. While the final pedestrian model is relatively simple in nature, it serves as a basis for the addition of larger amounts of data, a higher dimension of parameters, and a more complex ABM environment.

\section{ABC Inference of an ABM}

A simple stochastic model was created in NetLogo to demonstrate the use of ABC within an ABM. The simulation consists of a stochastic process with seven nodes arranged in a fork pattern (described in Section 4). This is taken to be a pedestrian network where nodes represent intersections and connections between the nodes are paths. All agents begin at the tail node 6 and travel to a connected node at each step. Observed data is first collected by setting all turning parameters to $\tfrac{1}{3}$ with the exception of arriving at Node 0 from 3. The probability of travelling vertically to 0 is set to 0.9. When this direction is taken, the action is denoted as $P(3^{\leftarrow 5}_{\rightarrow 0})$. That is, when the agent has travelled from 5 to 3, it then travels to 0. 

\begin{figure}[H]
    \centering
    \includegraphics[width=5cm]{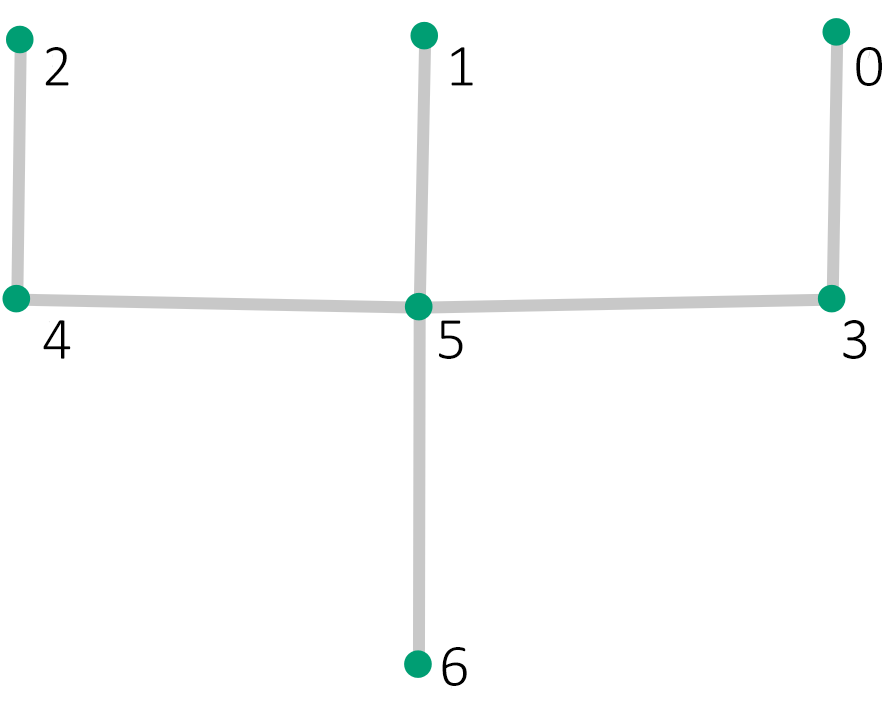}
    \caption{Directed graph of the Toy CBD model}
\end{figure}

Observed data was taken as an average of counts at Node 0 over ten runs. The known parameters are
\begin{align}
N(\text{Pedestrians}) = 15 \\
P(3^{\leftarrow 5}_{\rightarrow 0}) = 0.9
\end{align}
In reality, the number of pedestrians in any given system is often not known. While the proportional differences between the node counts depends on the directional probabilities, the magnitude of the counts is dependent on the number of pedestrians in the system. The observed data was an average of the count at node zero over ten runs, calculated as $x$. The prior parameters are
\begin{align}
P(3^{\leftarrow 5}_{\rightarrow 0}) \sim Beta(2, 2) \\
N(\text{Pedestrians}) \sim Unif(5,20)
\end{align}
The Beta distribution is a family of continuous probability distributions defined over the interval [0,1]. It is characterised by two parameters $\alpha$ and $\beta$ which control the shape. The expected value and variance of the distribution are defined as
\begin{align}
E[X] = \frac{a}{a+b}, \quad Var[X] = \frac{ab}{(a+b)^{2}(a+b+1)}
\end{align}
It is commonly used as a prior in Bayesian statistics to describe the probability of an event. It is the conjugate prior probability distribution for the Bernoulli, Binomial, Negative Binomial and Geometric distributions. A flat prior can be defined as $Beta(1,1)$. In the following example a prior of $Beta(2,2)$ was used as a wide prior, assuming little knowledge about the true probability. The expected value and variance of the Beta prior is given as
\begin{align}
E[X] &= \frac{a}{a+b} = \frac{2}{4} = \frac{1}{2} \\
Var[X] &= \frac{ab}{(a+b)^{2}(a+b+1)} = \frac{4}{(4)^{2}(5)} = 0.05
\end{align}

\begin{figure}[H]
    \centering
    \includegraphics[width=7cm]{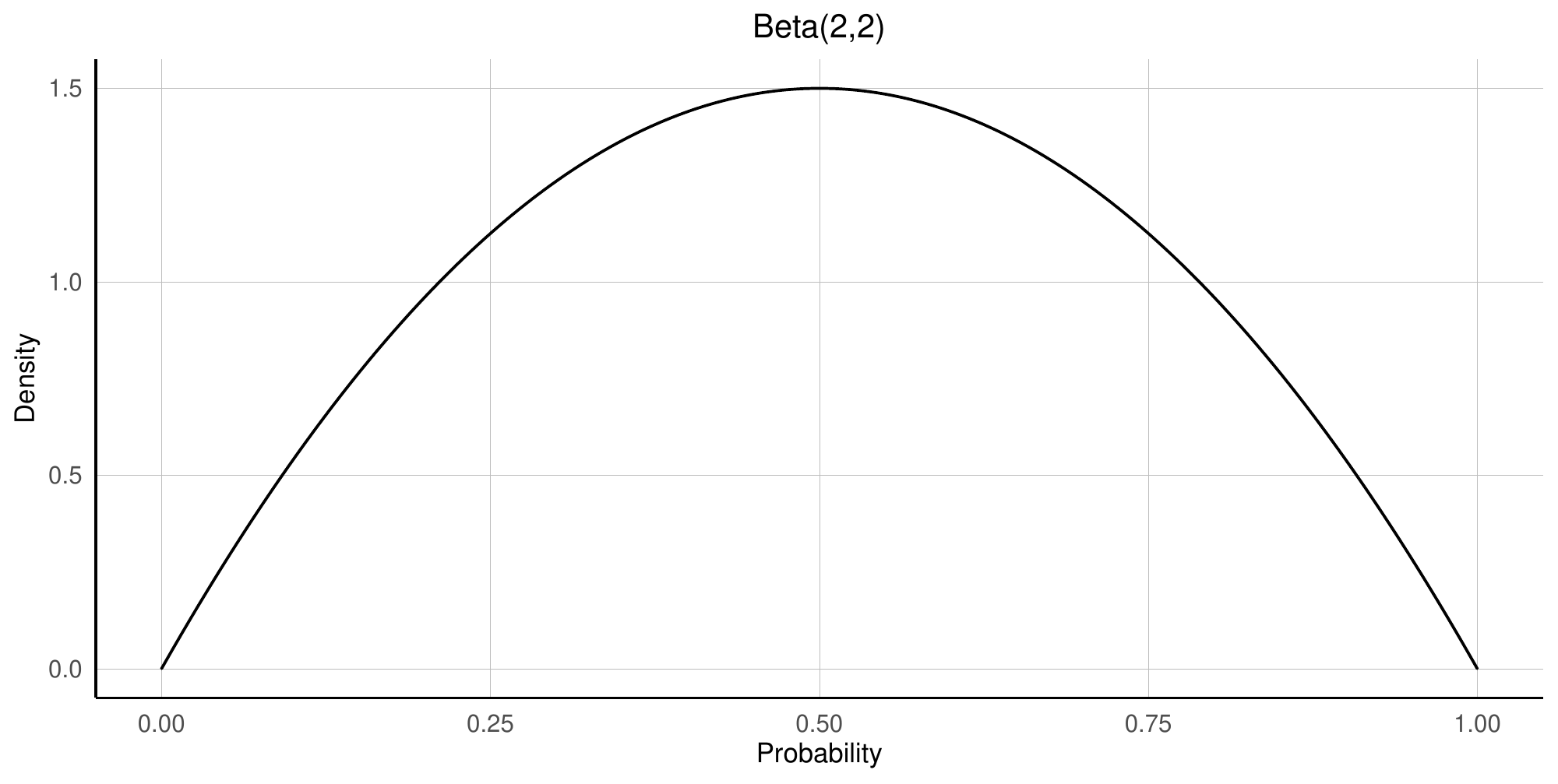}
    \caption{The Beta(2,2) prior distribution for the Toy Model.}
\end{figure}

Both Rejection ABC and ABC-SMC were used to infer the Beta parameter. The following algorithms were used:
\begin{algorithm}[H]\caption{ABC Rejection - Toy Model}\label{abc-rej}
\begin{algorithmic}[1]
\State Compute $\frac{1}{n}\sum_{i = 1}^{10}y(100)$, where $y$ is the count at Node 0 after 100 simulation ticks
\State Define a Euclidean distance measure $d(s(y), s(\hat{y})) = (s(y) - s(\hat{y}))^{2}$ 
\State Define a Beta prior distribution $\theta$ as $P(3^{\leftarrow 5}_{\rightarrow 0}) \sim Beta(2, 2)$
\For  {$i=1$ to $1000$} 
\State Sample a parameter from the prior distribution $\theta$
\State Simulate a dataset given the sampled parameter: $y^{*} \sim \pi(y^{*}|\theta^{*})$
\State Compute a summary statistic $s(\hat{y})$ for the simulated data
\State Add $d(s(y), s(\hat{y}))$ and $\theta_{i}$ to the distances vector
\EndFor
\State Find $\hat{\theta}$ as the top 1\% of the distances vector by distance
\State Approximate the posterior distribution $\theta$ from the distribution of accepted parameter values $\hat{\theta}$
\end{algorithmic}
\end{algorithm}

\begin{algorithm}[H]\caption{ABC-SMC}\label{abc-smc}
\begin{algorithmic}[1]
\State Compute $\frac{1}{n}\sum_{i = 1}^{10}y(100)$, where $y$ is the count at Node 0 after 100 simulation ticks
\State Set $\epsilon_{1} = 20$ and $t = 1$
\State Define a Beta prior distribution $\theta$ as $P(3^{\leftarrow 5}_{\rightarrow 0}) \sim Beta(2, 2)$
\State Define a Euclidean distance measure $d(s(y), s(\hat{y})) = (s(y) - s(\hat{y}))^{2}$ 
\For  {$i=1$ to $N$} 
\State Simulate $\theta \sim p(\theta)$
\State  Simulate $x \sim p(x|\theta)$ until $p(x,x_{obs}) < \epsilon_{1}$
\State Set $w_{i} = 1/N$
\EndFor
\For  {$t=2$ to $T$} 
\State Set $\epsilon_{t}$ as $0.75 \times \frac{1}{N} \sum_{i=1}^{N}t_{i}$
\State Calculate the weighted mean of the previous parameters as $\theta = \sum_{i=1}^{N}\theta^{(1)}_{i} w_{i}$
\For  {$i=1$ to $N$} 
\State Repeat until $p(x,x_{obs}) < \epsilon_{t}$:
\Indent
\State Pick $\theta$ from the $\theta^{(t-1)}$'s with probabilities $v_{j}^{(t-1)}$,
\State  Simulate $x \sim p(x|\theta)$
\EndIndent
\State Compute new weights as
\State \begin{align}
w_{i}^{(t)} \propto \frac{p(\theta_{i}^{(t)})}{\sum_{j}v_{j}^{(t-1)}K_{\theta,t}(\theta_{i}^{(t)}|\theta_{j}^{(t-1)})}
\end{align}
\EndFor
\State Normalise $w_{i}^{(t)}$ over $i = 1,...,N$
\EndFor

\end{algorithmic}
\end{algorithm}

\begin{table}[H]
\centering
\begin{tabular}{@{\extracolsep{4pt}}llccccccc}
\toprule   
 Prior $g(\theta)$ & Observed Data & Summary Statistic \\ 
\midrule
$\!\begin{aligned}[t]
    &\theta = Beta(2,2) \\
    \end{aligned}$ & $\{C_{i}(0)\}, i = 1,...,10$ & $\{\frac{1}{10}\sum_{i=1}^{10}C_{i}(0)\}$ \\ 
\bottomrule
\end{tabular}
\caption{Setup of the first Toy CBD model} 
\end{table}

The NetLogo simulation was controlled using the pyNetLogo library in Python. The Rejection ABC algorithm was programmed in Python.

\begin{figure}[H]
    \centering
    \includegraphics[width=10cm]{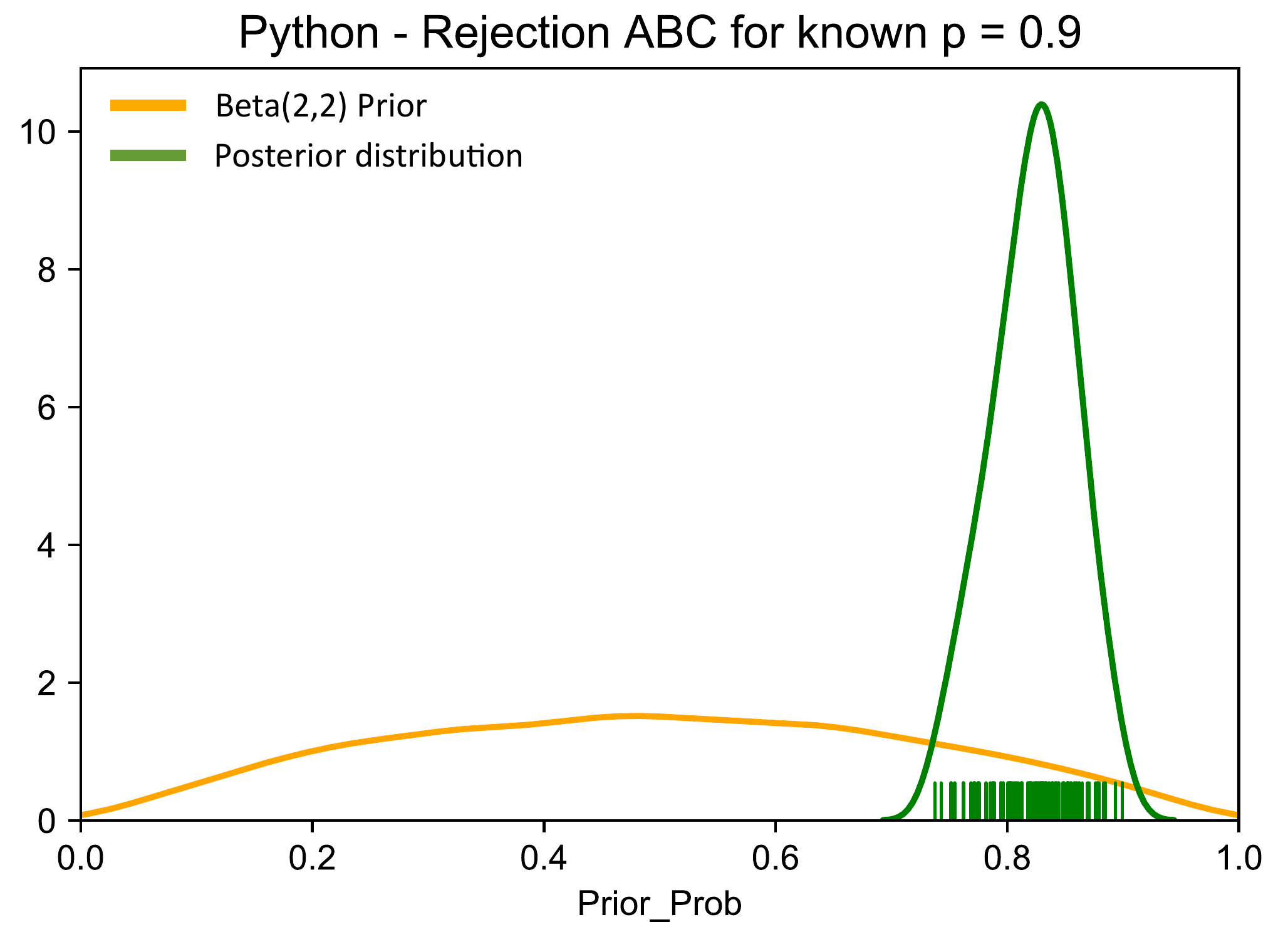}
    \caption{Posterior distribution (green) after 1000 simulations. The Prior Beta distribution is shown in orange.}
\end{figure}

\begin{table}[H]
\centering
\begin{tabular}{@{\extracolsep{4pt}}llccccccc}
\toprule   
 Simulations & Accepted Samples & Mean($\hat{\theta}$) & Running Time (mins) \\ 
\midrule
10000  & 1\% & 0.92 & 32.2 \\ 
\bottomrule
\end{tabular}
\caption{Results of ABC simulation} 
\end{table}
The posterior distribution appears to converge on the true parameter as epsilon decreases. However the running time is long, and 99\% of samples are rejected. The same model was run using ABC-SMC to assess the difference in efficiency. The settings of the model are
\begin{table}[H]
\centering
\begin{tabular}{@{\extracolsep{4pt}}llccccccc}
\toprule   
 Populations & Population Size & Initial Epsilon\\ 
\midrule
5  & 250 & 10\\ 
\bottomrule
\end{tabular}
\caption{ABC-SMC Settings, Toy Model} 
\end{table}

\begin{figure}[H]
    \centering
    \includegraphics[width=15cm]{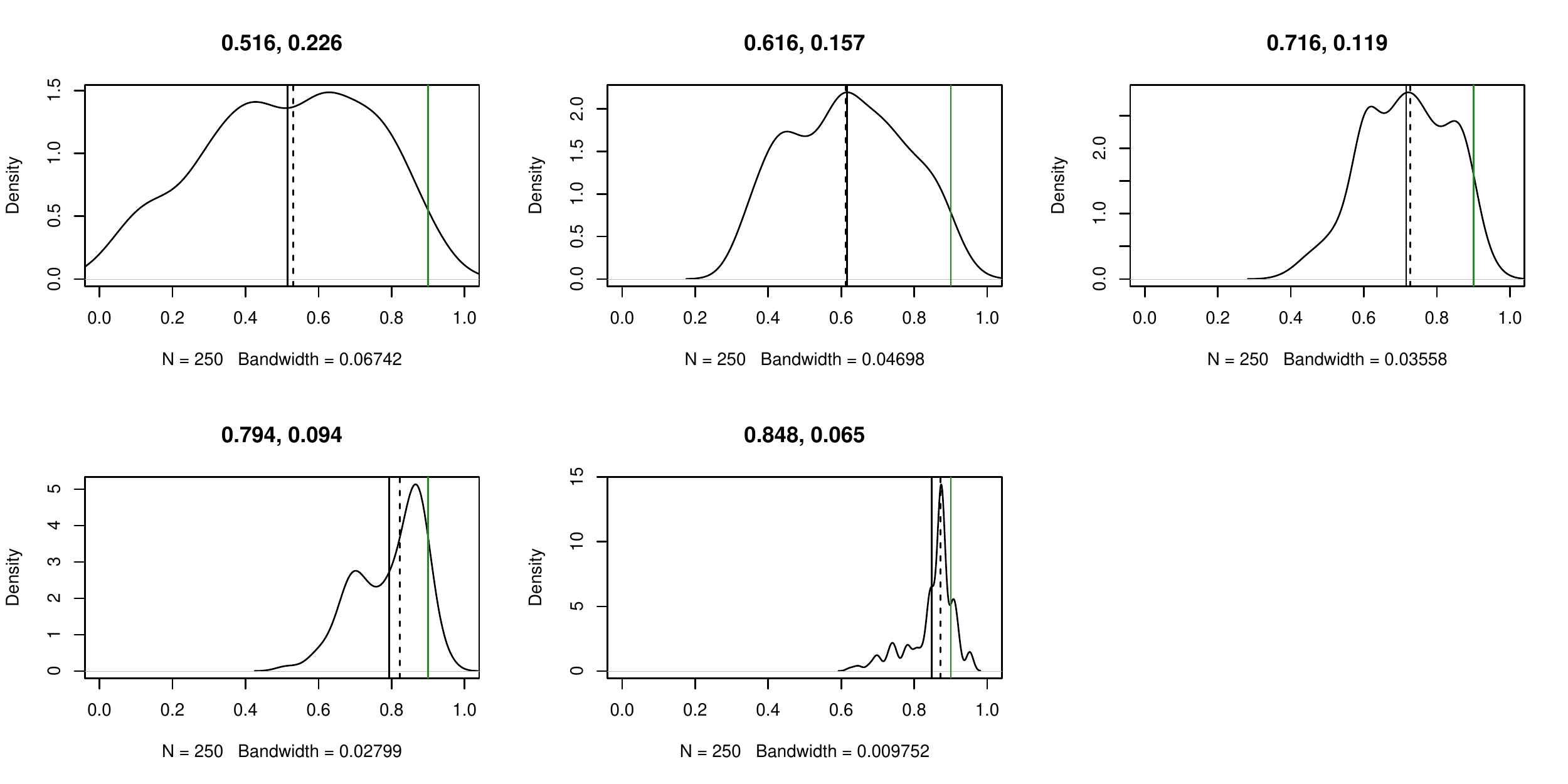}
    \caption[Progression in the posterior mean with a known $\mu$ = 0.9.]{Progression in the posterior mean with a known $\mu$ = 0.9. The true parameter is shown in green, the posterior mean in black, and the posterior median by a dashed line.}
\end{figure}

\begin{table}[H]
\centering
\begin{tabular}{@{\extracolsep{4pt}}llccccccc}
\toprule   
 Simulations & Accepted Samples & Mean($\hat{\theta}$) & Running Time (mins) \\ 
\midrule
1542  & 250 & 0.848 & 10.7\\ 
\bottomrule
\end{tabular}
\caption{Results of ABC simulation} 
\end{table}

Here the acceptance rate is much higher (6\%). The algorithm converges on the posterior much more quickly than the rejection algorithm. \\

\textbf{Extended Network Model}\\

This model can be extended to inferring parameters for a larger set of nodes. For nodes with multiple possible directions, the Dirichlet distribution can be used as a multivariate generalisation of the Beta prior. The observed data can also be extended to multiple dimensions, where counts are observed at each node in the network. Consider a network with $N$ nodes, each with $K$ edges extending from the node, and $D$ observed counts. The model can be setup as:
\begin{align}
\theta &= \theta_{1},...,\theta_{N}, \text{where}\;\theta_{i} \sim Dirichlet(\alpha_{1},...,\alpha_{K}) \\
Y &= \{C_{i}\}, i = 1,..,N
\end{align}
This model will serve as a framework for modelling the movement of pedestrians in such a network. \\

\textbf{Pedestrian Model Parameters}\\
An intuitive way to describe a pedestrian making a decision for a direction to travel at each intersection is that of a Categorical distribution. Given $N$ possible directions, there is a set of probabilities $P$ for travelling in each direction. The Categorical distribution is a generalisation of the Bernoulli distribution with probability $p$ of outcome 1 and $q = 1-p$ of outcome 2. The generalisation extends the model to $K$ possible categories. The probability of each category or outcome is specified with $p_{1},...,p_{k}$ (Categorical distribution, n.d.).\\

In Bayesian statistics, the Dirichlet distribution is the conjugate prior for the Categorical distribution. In a model where each data point follows a Categorical distribution with an unknown parameter vector \textit{p}, this is treated as a random variable $X$ with a prior Dirichlet distribution. The posterior distribution of the parameter is also Dirichlet. Knowledge of a parameter can be updated by incorporating observations one at a time, following a formally defined mathematical process. The expected value of the posterior distribution is
\begin{align}
E[p_{i} | \mathbb{X}, \alpha] = \frac{c_{i} + \alpha{i}}{N + \sum_{k}\alpha_{k}}
\end{align}
The expected probability of seeing a category $i$ is equal to the proportion of successful occurrences of $i$ out of all total occurrences. \\

\textbf{Dirichlet Distribution} \\
If there are three possible directions at a given intersection $I$, equal probabilities of each direction can be specified by setting each alpha to the same value, giving a symmetric Dirichlet distribution. The size of each alpha specifies the variance. For example, setting the prior distribution $\theta \sim Dirichlet(2,2,2)$ gives a mean and variance of:
\begin{align}
E[X_{i}] &= \frac{\alpha_{i}}{\alpha_{0}}, \alpha_{0} = \sum^{K}_{i=1}\alpha_{i} \\
E[X_{2}] &= \frac{2}{6} = \frac{1}{3} \\
Var(X_{i}) &= \frac{\alpha_{i}(\alpha_{0}-\alpha_{i})}{\alpha^{2}_{0}(\alpha_{0}+1)} = \frac{2(6-2)}{36(6+1)} = 0.0317
\end{align}

\section{Hamilton CBD Pedestrian Model}

\textbf{Research Questions} \\
Traffic modelling is an important aspect of transport engineering. Modelling transport networks aids in understanding hidden phenomena of travel and allows for the testing of infrastructure design. There is a need in modern urban environments to prioritise sustainable modes of transport, of which walking plays a key role. \\

A pedestrian model can be viewed in the context of an Agent Based Model. At each \textit{tick}, pedestrians make a decision to move to a new location using a complex internal thought process. This decision is based on attributes including, but not limited to; the desired destination, structure of the network and characteristics of the built environment. \\

Understanding pedestrian flows is vital is designing an environment for pedestrians such that they reach their destination in a safe and enjoyable manner. To aid in this understanding, data first needs to be collected on where pedestrians are travelling in the network. \\

Pedestrian counters have been setup in the Hamilton CBD at 21 key locations. These counters are situated on store verandahs and use 3D StereoVision cameras, giving them the ability to distinguish the direction a pedestrian is travelling and provide distance and height measurements. Each camera records a total count in two directions in five minute intervals. Currently there is no extensive analysis being conducted with the data from these sensors, apart from tracking the short term and long term trend. Beyond the 21 sensors operated by HCC, there is a long term vision of extending this to more locations to better understand pedestrian flow. The model described below serves as an initial exploration of how a pedestrian model can be developed for Hamilton using count data. The model will aim to answer a key question of pedestrian flows:\\

\textit{What are the probabilities of turning in each possible direction, for each intersection in the Hamilton CBD?} \\

These probabilities are hidden parameters in the network, which result in the counts observed at the pedestrian counters. Using the observed data, we can infer these probabilities parameters using an Agent Based Model and ABC-SMC.

\begin{figure}[H]
    \centering
    \includegraphics[width=5cm]{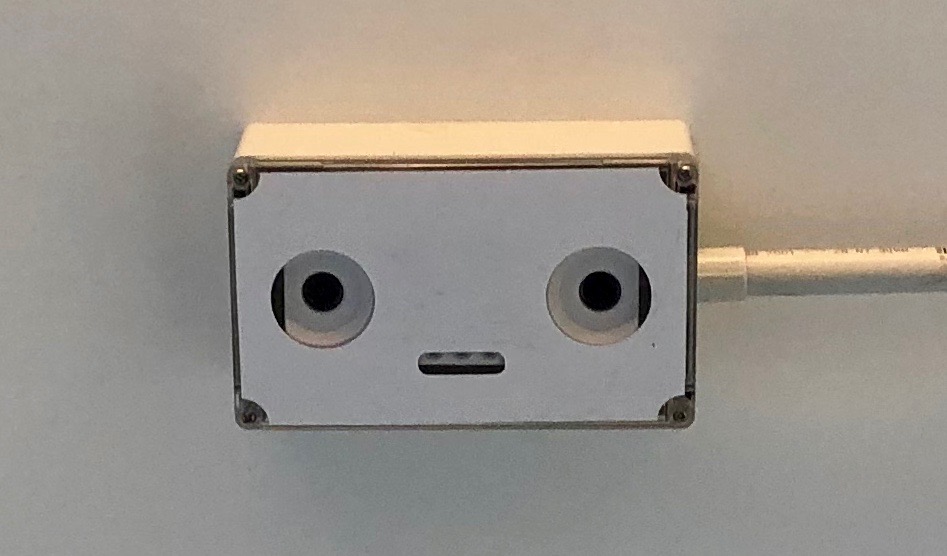}
    \caption{A pedestrian counter located on the verandah outside Crate Clothing, Ward St}
\end{figure}

The model focuses on the four pedestrian counters near the Centre Place shopping centre at the intersection of Ward St and Worley Place. The system is defined as a network with entry/exit points, intersections nodes, and counting nodes. The network is loaded from a shapefile of the Hamilton City walking network (NZTM) and is geospatially consistent with the real world location. The 4 pedestrian counters are included as a certain type of node which counts travel through itself. The additional counters (denoted with a dashed circle below) are not installed counters. However for the purposes of covering all entry and exit points they are included as such, with mock data created according to a best guess. The data from these counters is used to validate the ABM. \\

For simplicity purposes, the network was simplified to two intersections and four counters. The entry to the northern Centre Place building was ignored, but could be added in future models in addition to extending the network to a larger portion of the CBD when the data is available.

\begin{figure}[H]
    \centering
    \includegraphics[width=9cm]{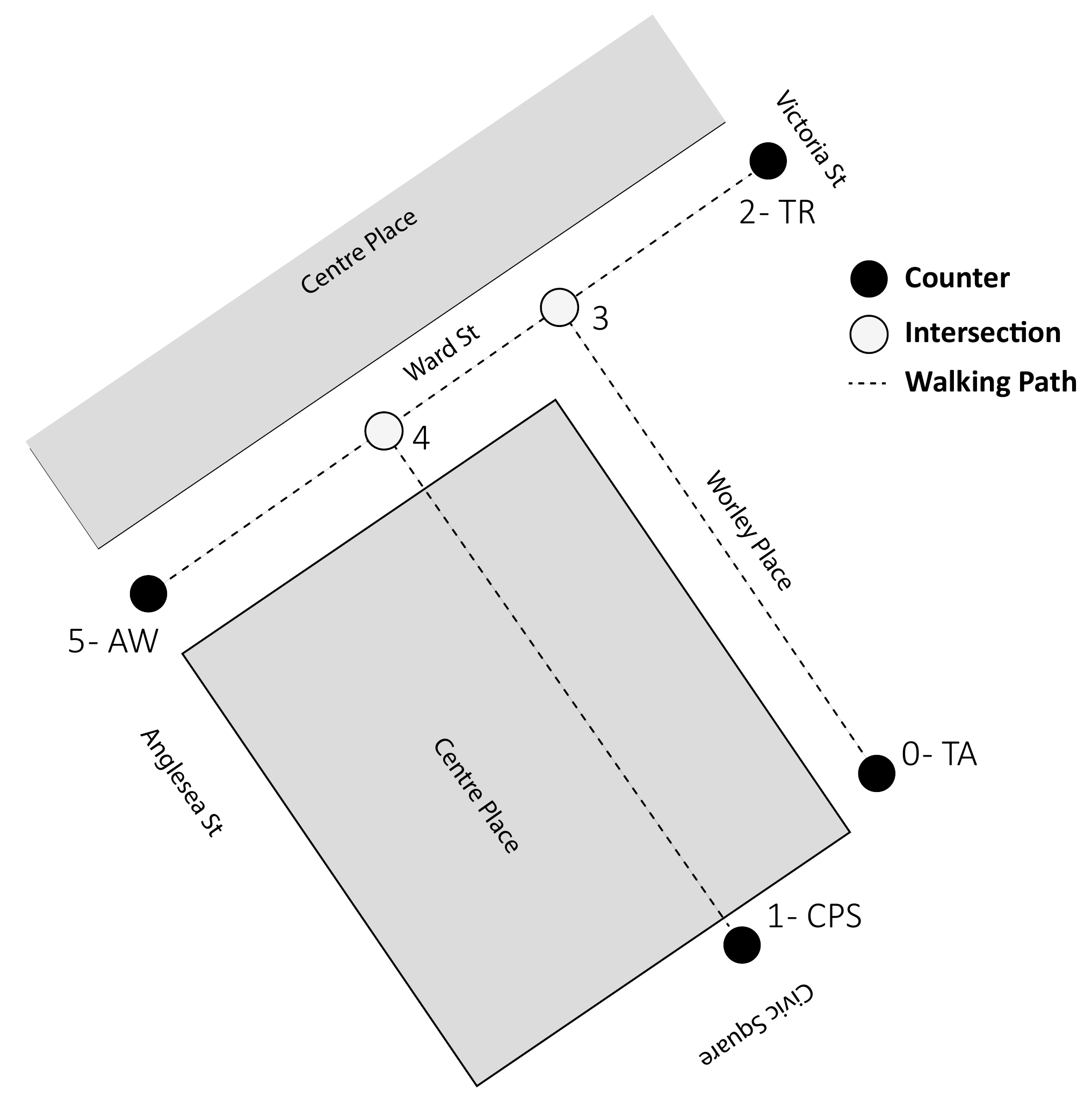}
    \caption{The Centre Place walking network.}
\end{figure}

An Agent Based Model was developed in NetLogo to simulate the counts at the four counters, with validation carried out using an ABC-SMC with Adaptive Weights model. The data for the four existing counters was first explored to understand how the count varies over time at each location.

\begin{table}[H]
\centering
\begin{tabular}{rrlll}
  \hline
Count & Location.Direction & Day & Time \\ 
  \hline
14 & Anglesea Ward NE & 2018-07-07 & 09:00:00 \\ 
36 & Anglesea Ward NE & 2018-07-07 & 10:00:00 \\ 
40 & Anglesea Ward NE & 2018-07-07 & 11:00:00 \\ 
56 & Anglesea Ward NE & 2018-07-07 & 12:00:00 \\ 
   \hline
\end{tabular}
\caption{An example of the formatted data from the pedestrian counters}
\end{table}

\begin{figure}[H]
    \centering
    \includegraphics[width=15cm]{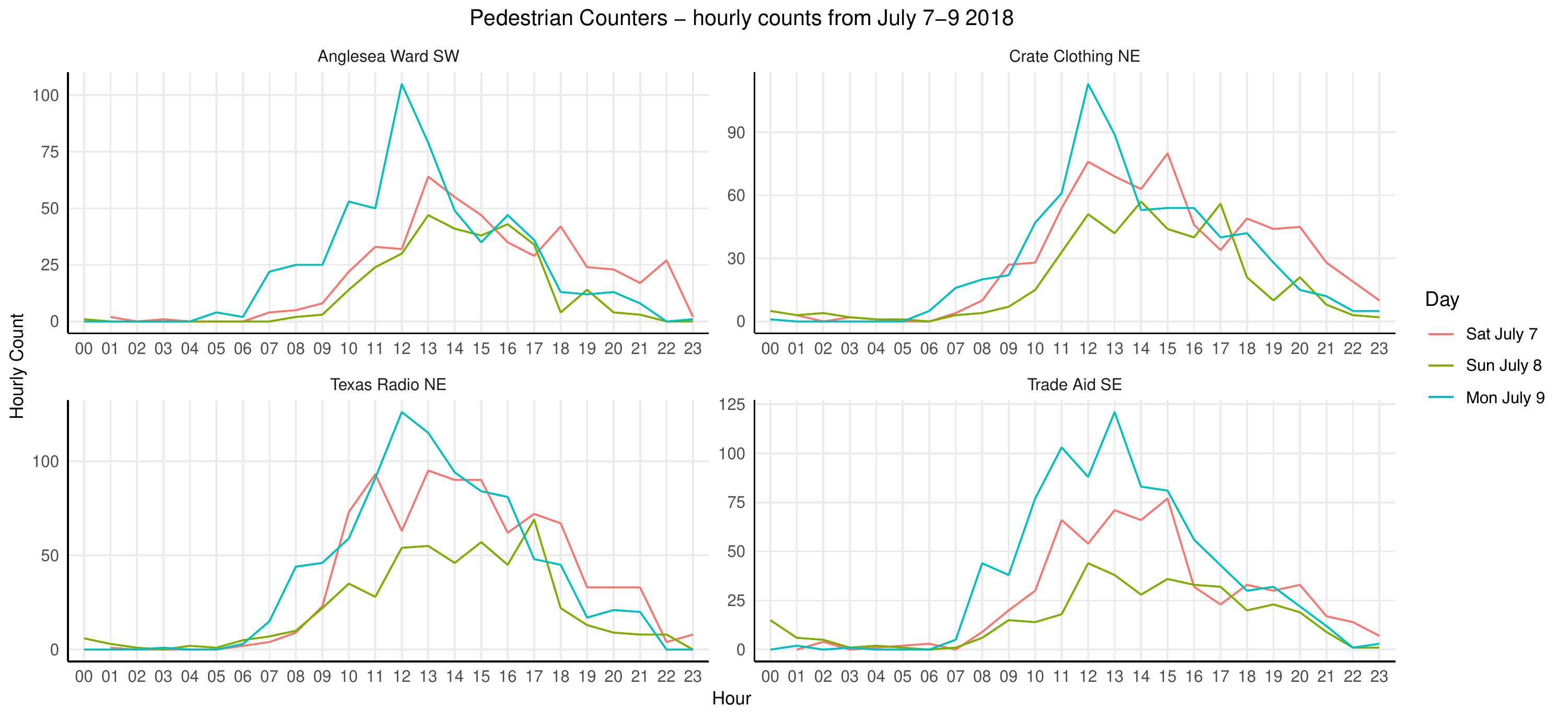}
    \caption{The pedestrian count from July 7-9, 2018 for the four observed counts to be used in the model}
\end{figure}
For each counter, the count changes significantly over the course of a day. The peak is generally around lunchtime. There is a slight increase in the evening before falling to a minimum level during the early morning hours. All four counters appear to vary significantly from the weekend to Monday. The count during the lunchtime peak was investigated further.
\begin{figure}[H]
    \centering
    \includegraphics[width=12cm]{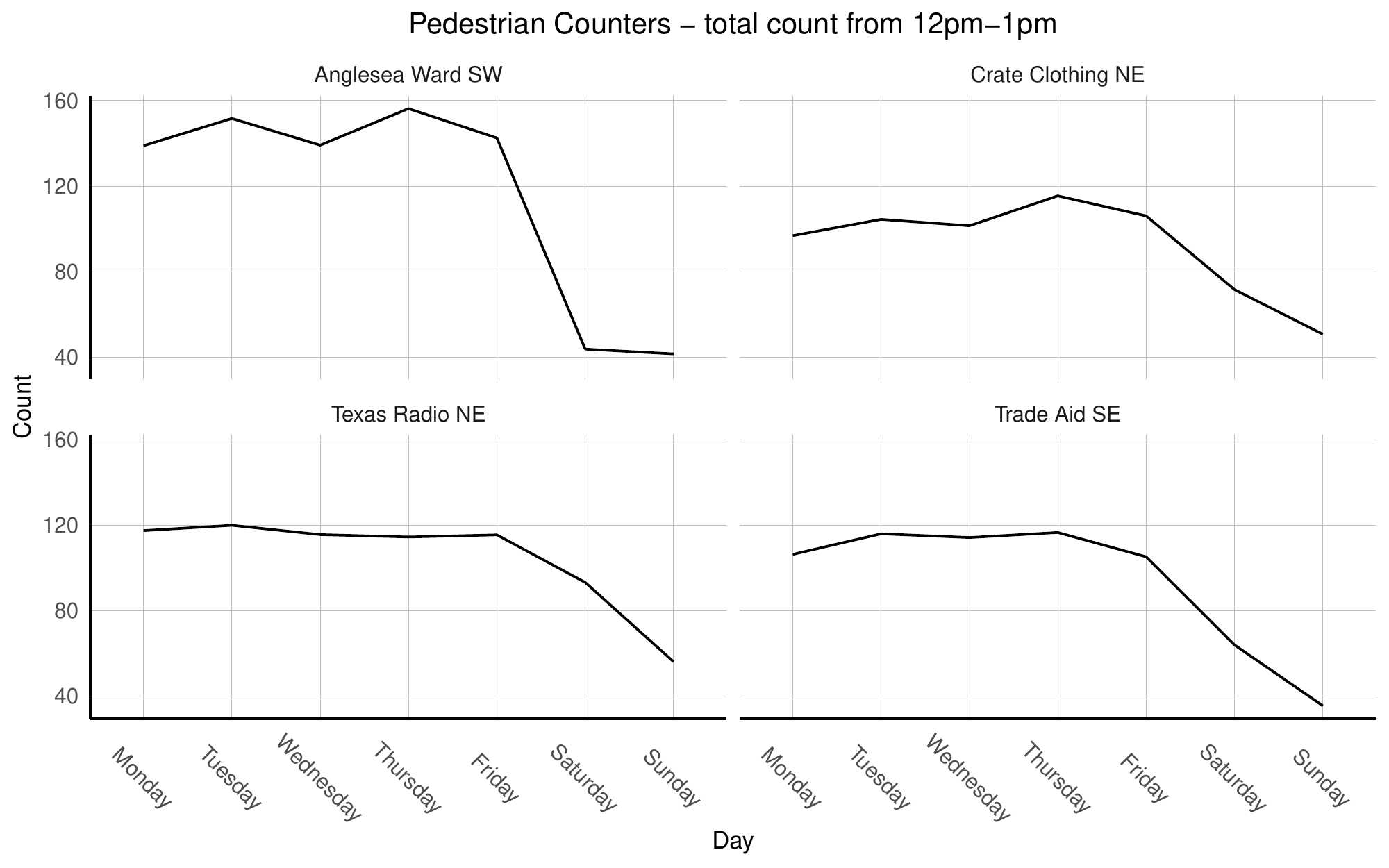}
    \caption{The average pedestrian count from 12pm-1pm for the four counters to be used in the model}
\end{figure}

All four locations appear to have similar counts over the course of a working week. There is a decrease in the count during Saturday and Sunday, particularly at the Anglesea/Ward counter. Sunday experiences the lowest counts for the week. In terms of magnitude, the Anglesea/Ward counter has the highest foot traffic. 12pm-1pm is a key time to understand pedestrian flows, as many people working in the CBD leave their place of work to eat or shop. Understanding the movement of pedestrians at this time will provide valuable insight into where people choose to travel.

\subsection{ABM Design}

The pedestrian network is represented in NetLogo as a series of nodes and connecting links. This can be viewed as a undirected, acyclic graph with six nodes and five edges. It is not possible to travel back to a node which an agent previously travelled to. The agents (pedestrians) travel between nodes via the links. At each tick of the simulation, the agents choose to travel to a connected node using a set of probabilities. The probabilities at nodes $\{3,4\}$ are the parameters of the ABM which will influence the counts at each exit node $\{5,2,1,0\}$. They are modelled using two symmetric Dirichlet distributions $\sim Dirichlet(t, k)$, where $t$ is the number of categories and $k$ is the concentration hyperparameter, set to be equal for all $t$. For two distributions and three categories this is denoted as:
\begin{align}
\theta_{1} \sim Dirichlet(3, 3, 3) \\
\theta_{2} \sim Dirichlet(3, 3, 3)
\end{align}

\begin{enumerate}
\item{\textbf{Purpose}} \\
To create a valid simulation of the pedestrian flows around the Ward/Worley intersection using observed data.

\item{\textbf{Entities, State Variables and Scales}} \\
The single agents in the network are pedestrians. They have two variables - their current node and a goal node. They travel discretely from one node to another.

\item{\textbf{Process overview and scheduling}} \\
At each tick of the simulation, each pedestrian in the network makes a decision to move to a certain node, and then moves to that node. These movements occur simultaneously.

\item{\textbf{Design Concepts}} \\
There is no interaction between pedestrians. The agents will exit and enter the system according to a survival function.

\item{\textbf{Initialisation}} \\
There are initially a number of agents assigned to entry/exit nodes, with probabilities of assignment proportional to the entry counts at the nodes.

\item{\textbf{Input Data}} \\
The input data is the observed entry counts for each node, used to assign agents to entry/exit nodes at the start of the simulation.
\end{enumerate}

The agents do not return to the node which they came from. There are two possible nodes to travel to at each intersection. Formally, the three directions at an intersection $n$ can be represented as $\{d_{1},d_{2},d_{3}\}$. The corresponding nodes if each respective direction is chosen are $\{n_{1},n_{2},n_{3}\}$. The probability of turning back is eliminated, and hence the remaining probabilities are normalised by dividing each by the sum.

\subsection{Initial Model Analysis}

The simulation is controlled over multiple runs with RNetLogo. Through this package, variables can be sent to and from R. A demonstration is shown below with a simple change in probabilities. The first run is conducted with even probabilities of turning in all directions. The probabilities are then modified with the intention to increase counts in the SW direction at Anglesea/Ward. This is done by changing the probability of a turn towards the Anglesea/Ward node to 0.95. Other possible directions $n$ are set to $\tfrac{0.05}{n}$. The mean count is taken over 10 runs of each simulation.

\begin{figure}[H]
    \centering
    \includegraphics[width=12cm]{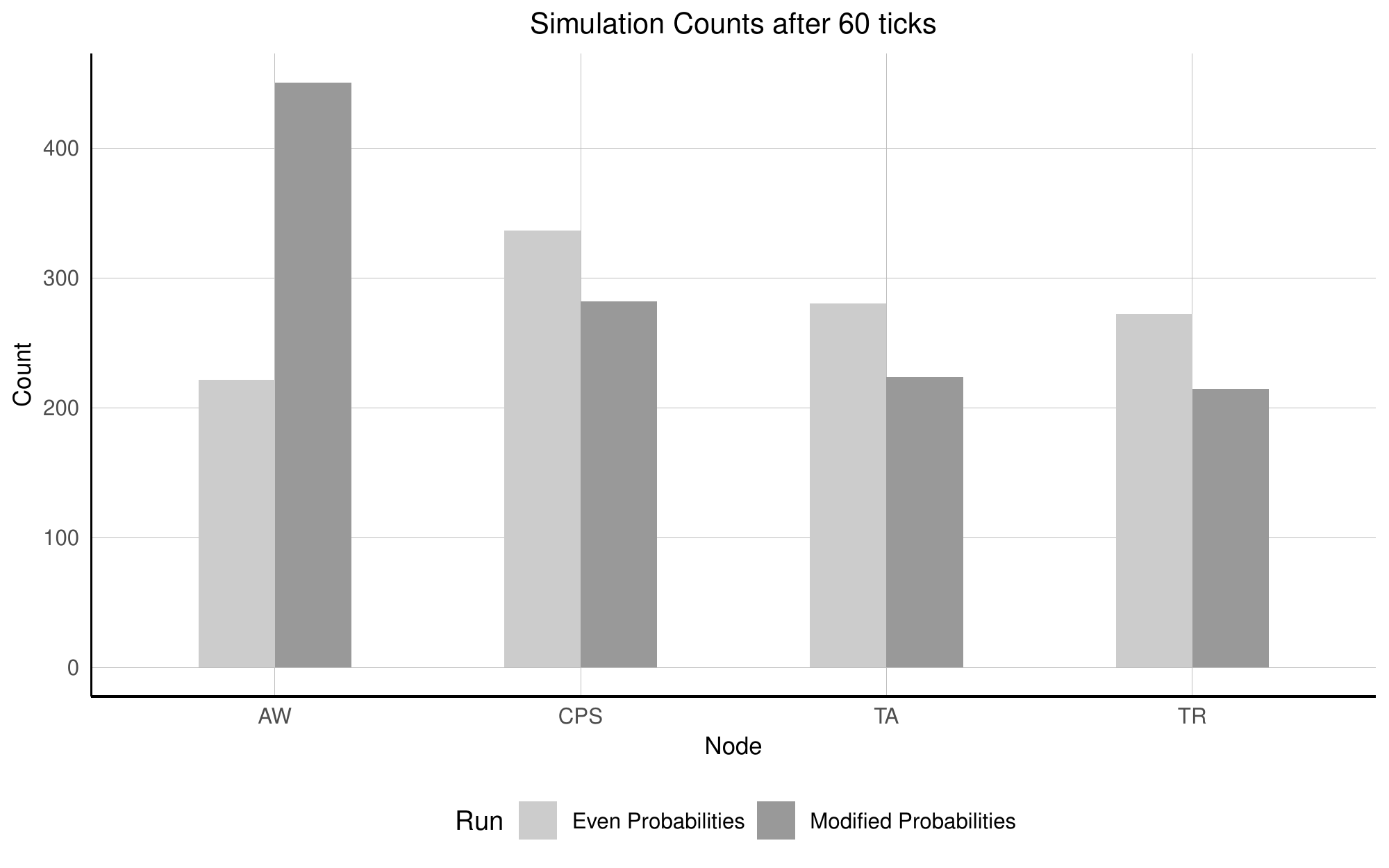}
    \caption{Changes in simulated counts after modification of turning probabilities.}
\end{figure}

As shown above, increasing the probability of turning SW to Anglesea/Ward approximately doubles the count across the node, while the count for the three other nodes decreases slightly. \\

\subsection{Observed Data}

The first CBD model will aim to create a validated ABM to accurately simulate the pedestrian movements for a typical weekday at 12pm. To gather the observed data, two months of hourly counts were extracted from August 1 - September 31. The data was subset to weekdays at 12pm and taken as the observed data. The mean of each is counter was set as the summary statistic. \\

For the counters either not installed or without data available, mock data was created according to a best guess. As a result the output from the model does not completely represent the real world system, but remains a valid framework for when the observed data is available. The process to create the mock data is described in the table below. The count from an existing counter is denoted by $C_{x}$, where $x$ is the name given to the counter.

\begin{table}[H]
\centering
\begin{tabular}{@{\extracolsep{4pt}}llccccccc}
\toprule   
 Node & Function & Count\\ 
\midrule
5 & $C_{AW} \times 2.2$ & 321 \\ 
2 & $C_{TR} + C_{CC}$ & 222 \\ 
0 & $C_{TA} \times 1.8$ & 202 \\ 
1 & $380$ & 380 \\ 
\bottomrule
\end{tabular}
\caption{Creation of the observed data for the four counters} 
\end{table}

\subsection{Model Design}

At each intersection in a walking network, pedestrians make a choice of a direction to turn. This can be described in general terms by the set $S = \{d_{i}\}, i = 1,...n$, where $n$ is the number of possible directions. The probabilities in this set will vary for each intersection. For simplicity purposes, the backwards direction will be ignored. A Dirichlet prior is used at each intersection. The Dirichlet distribution is a family of continuous multivariate probability distributions (J. Lin, 2016).
\begin{flalign}
f(x_{1},...,x_{K};\alpha_{1},...,\alpha_{K}) = \frac{1}{B(\alpha)}\prod_{i=1}^{K}x_{i}^{\alpha_{i}-1}, \\
B(\alpha) = \frac{\prod_{i=1}^{K}\Gamma(\alpha_{i})}{\Gamma(\sum_{i=1}^{K}\alpha_{i})}
\end{flalign}
The Dirichlet distribution is well suited as a Bayesian prior for a set of probabilities in $k$ choices. The number of parameters is equal to the number of possible directions. The probability of each direction is equal for input into the model, and is then inferred as a parameter using ABC. In total there are 6 nodes - 4 for recording counts and 2 non-recording intersections. Probability distributions are placed over the two intersections $\{3,4\}$. Intersections 3 and 4 in Figure 0.15 have 3 choices of direction. These two directions will vary depending on the node a pedestrian previously came from. \\

\begin{table}[H]
\centering
\begin{tabular}{@{\extracolsep{4pt}}llccccccc}
\toprule   
 Current Node & Previous Node & Next Nodes & Possible Directions\\ 
\midrule
3 & 2 & {0,4} & 2 \\ 
3 & 0 & {2,4} & 2 \\ 
3 & 4 & {0,2} & 2 \\ 
4 & 3 & {1,5} & 2 \\ 
4 & 1 & {3,5} & 2 \\ 
4 & 5 & {1,3} & 2 \\ 
\bottomrule
\end{tabular}
\caption{Possible directions for node travel} 
\end{table}

\begin{figure}[H]
    \centering
    \includegraphics[width=10cm]{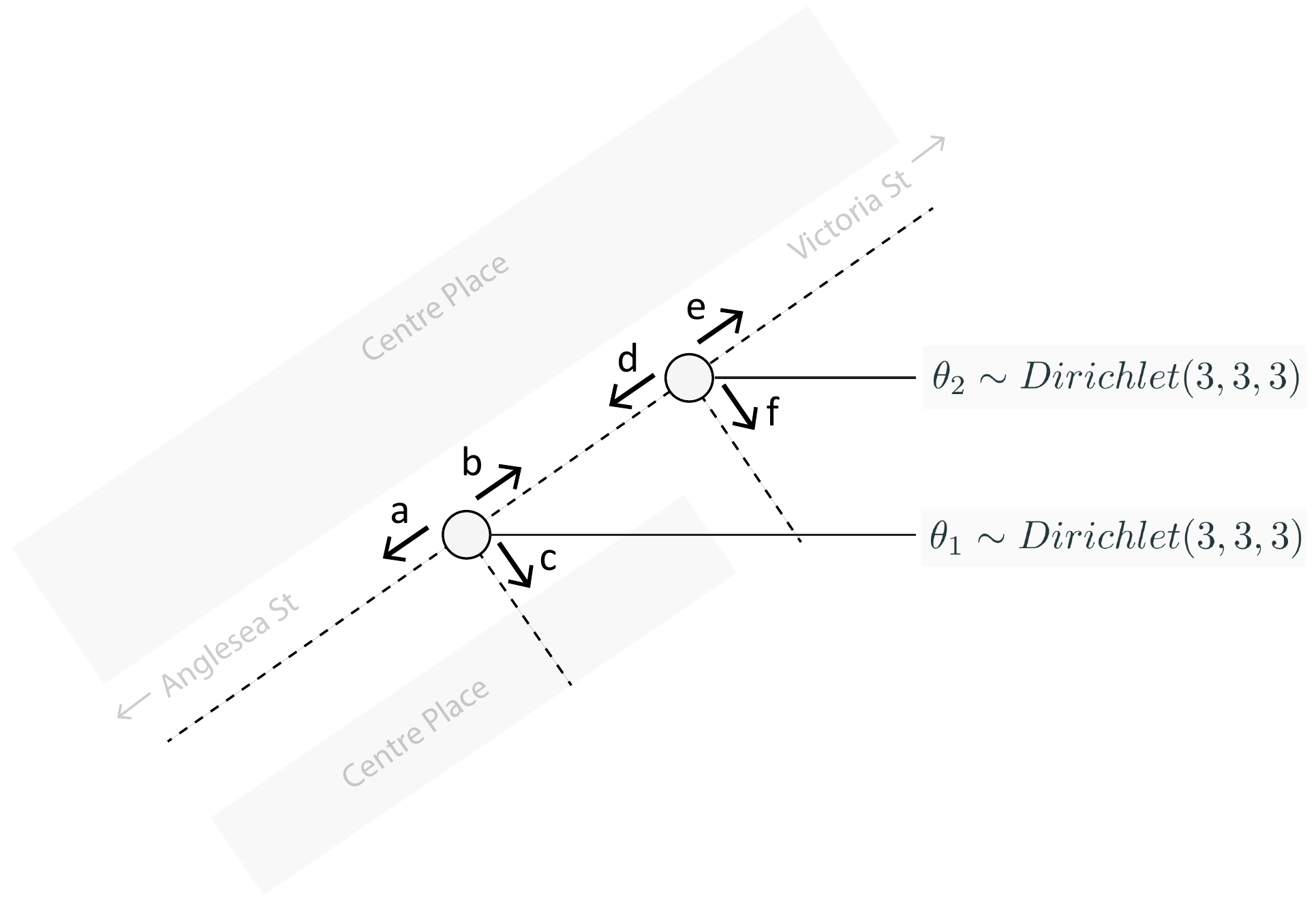}
    \caption{Prior Dirichlet distribution at the two intersections and the probability parameters a-f.}
\end{figure}

The aim of the agent based model with ABC validation is to understand the probabilities of a pedestrian turning at certain intersections. It is understood that there is a set of probability distributions which will produce simulated data close to that of the observed data. These probability parameters will assist in understanding the movement of pedestrians around the Centre Place shopping centre, and how this varies at different times of the day. The first model will infer these probabilities for 12pm on a typical weekday. It is expected that the foot traffic will be high during this time as a result of workers in the CBD using their lunch break to go out to eat, shop, or walk. \\

An additional parameter is the number of agents in the system over the hour span. This is defined with a Uniform prior, $\sim Uniform(a,b)$. $a$ and $b$ are chosen according to the observed entry count data and number of ticks in the simulation. The total entry over one hour from 12pm-1pm was calculated as 1021. For 60 ticks representing one hour, this gives an approximate entry count per minute of $\sim$17. However, the number of ticks in the simulation significantly affects the simulated counts. For example, if 100 ticks result in an observed count of 160, at 60 ticks the count will be approximately $\frac{100}{60} \times 160 \approx 267$. There is no single calculation for the number of pedestrians to use in the ABM, as it will be influenced by the design of each network. One method to find it is to set a wide Uniform prior for the number and infer the most likely value with ABC. This highlights a benefit of the ABC-SMC algorithm - even if the Uniform prior does not cover the true parameter, the sampling step will allow for parameters beyond the prior range. This is the method used in the Hamilton CBD pedestrian model. \\

\textbf{Dynamic Entry/Exit}

A more realistic method of simulating pedestrians is to begin with a given number in the system, and use a function at each tick to generate a certain number of agents to enter the system. A parameter $\alpha$ can be set to manage the entry rate such that the simulation closely matches the observed data. This method was not used in the Hamilton CBD model, although it will be investigated further for the extended model.

\subsection{ABC Model}
The ABC-SMC model is defined by four key measures.
\begin{enumerate}
\item Prior distributions of the parameters $\theta$. At each intersection a Dirichlet distribution is defined as a prior with $K$ equal to the number of possible directions.
\item A summary statistic calculated from the observed and simulated data. The output from each simulation will be counts in the exit direction at four nodes. The mean will be calculated for each of these four measures and compared against the mean counts from the observed data at 12pm on a weekday.
\item A distance measure D. This was chosen as Euclidean distance:
\begin{align}
d = \sqrt{\sum_{i}^{N}(x_{i} - x_{i(obs)})^{2}}
\end{align}
\item An initial epsilon $\epsilon$ as a threshold for the distance measure. This is dynamically set during the model simulation using a quantile of the previous population distances.
\end{enumerate}

To generate data $y$, the ABM will be run for 100 ticks. This will represent the pedestrian flow from 12pm-1pm on a weekday. Intersection nodes will be given prior probabilities of turning in each direction, sampled from $g(\theta)$. The generated data is the counts at each counting node. At the end of each simulation the total count at each node is used at the simulated data. The ABC algorithm is given below:

\begin{algorithm}[H]\caption{ABC-SMC}\label{abc-smc}
\begin{algorithmic}[1]
\State Compute the summary statistics as the mean count at 12pm, across all weekdays, for each of the four entry/exit nodes
\State Set $\epsilon_{1} = 350$ and $t = 1$
\State Set $\theta^{1} \sim Dirichlet(3,3,3)$, $\theta^{2} \sim Dirichlet(3,3,3)$ and $\theta^{3} \sim Unif(40,60)$
\State Define a Euclidean distance measure $d = \sqrt{\sum_{i}^{N}(x_{i} - x_{i(obs)})^{2}}$ 
\For  {$i=1$ to $N$} 
\State Simulate $\theta_{i}^{(1)} \sim p(\theta^{(1)}), \theta_{i}^{(2)} \sim p(\theta^{(2)}), \theta_{i}^{(3)} \sim p(\theta^{(3)})$
\State  Simulate $x \sim p(x|\theta_{i}^{(1)},\theta_{i}^{(2)}),\theta_{i}^{(3)})$ until $p(x,x_{obs}) < \epsilon_{1}$
\State Set $w_{i} = 1/N$
\EndFor
\For  {$t=2$ to $T$} 
\State Compute data based weights $v_{i}^{(t-1)} \propto w_{i}^{(t-1)}K_{x,t}(x_{obs}|x_{i}^{(t-1)})$
\State Set $\epsilon_{t}$ as $0.75 \times \frac{1}{N} \sum_{i=1}^{N}t_{i}$
\For  {$i=1$ to $N$} 
\State Repeat until $p(x,x_{obs}) < \epsilon_{t}$:
\Indent
\State Pick $\theta_{i}^{*}$ from the $\theta_{j}^{(t-1)}$'s with probabilities $v_{j}^{(t-1)}$
\State Pertube the sampled particle $\theta_{i}^{*}$
\State Simulate $\theta_{i}^{(1)} \sim p(\theta^{(1)}), \theta_{i}^{(2)} \sim p(\theta^{(2)}), \theta_{i}^{(3)} \sim p(\theta^{(3)})$
\State  Simulate $x \sim p(x|\theta_{i}^{(1)},\theta_{i}^{(2)},\theta_{i}^{(3)})$
\EndIndent
\State Compute new weights as $w_{i}^{(t)} \propto \frac{p(\theta_{i}^{(t)})}{\sum_{j}v_{j}^{(t-1)}K_{\theta,t}(\theta_{i}^{(t)}|\theta_{j}^{(t-1)})}$
\EndFor
\State Normalise $w_{i}^{(t)}$ over $i = 1,...,N$
\EndFor

\end{algorithmic}
\end{algorithm}

\subsection{Agent Behaviour}

At creation of the model a given number of pedestrians $P$ are generated. This number is given in the first ABC population by a wide Uniform prior. By pertubing subsequent particles from the population, the number will converge to the true posterior even if the initial prior does not cover the true value. The agents are assigned to exit nodes according to probabilities proportional to the entry counts at 12pm on a typical weekday:

\begin{align}
  P(I) = \frac{C_{I}}{\sum_{i=1}^{4} C_{i}}, i \in \{0,1,2,5\},
  \end{align}
where $C_{x}$ is the count at node $x$. \\

At each simulation tick, the agents individually make a decision of an adjacent node to move to using the set of probabilities from the ABC model. At each ABC iteration $1,...N$, sampled probabilities are sent to the ABM and the simulation is run for 100 ticks to represent one hour. When an agent moves to an entry/exit node, the count for that node is updated. The total count at each node after 100 ticks is reported back to the ABC model as the simulated data.

\subsection{Results}

\begin{table}[H]
\centering
\begin{tabular}{@{\extracolsep{4pt}}llccccccc}
\toprule   
 A & B & C & D & E & F & Num Peds \\
\midrule
0.488 & 0.202 & 0.310 & 0.380 & 0.320 & 0.300 & 27.408 \\
\bottomrule
\end{tabular}
\caption{Posterior Means} 
\end{table}

\begin{table}[H]
\vspace*{-0.5 cm}
\centering
\begin{tabular}{@{\extracolsep{4pt}}llccccccc}
\toprule   
 Populations & Simulations & Accepted Samples & Running Time (mins) \\ 
\midrule
10 & 7620 & 3000 & 282 mins \\ 
\bottomrule
\end{tabular}
\caption{Results of ABC simulation} 
\end{table}

\begin{table}[H]
\vspace*{-0.5 cm}
\centering
\begin{tabular}{@{\extracolsep{4pt}}llll}
\toprule   
 Counter & Observed Count & Simulated Count & Difference \\ 
\midrule
AW & 321 & 306.041 & 14.959\\ 
TR & 222 & 224.2908 & -2.2908\\ 
TA & 202 & 214.0074 & -12.0074\\ 
CPS & 380 & 393.1164 & -13.1164\\ 
\bottomrule
\end{tabular}
\caption{Results of ABC simulation} 
\end{table}

\begin{figure}[H]
  \centering
    \includegraphics[width=14cm]{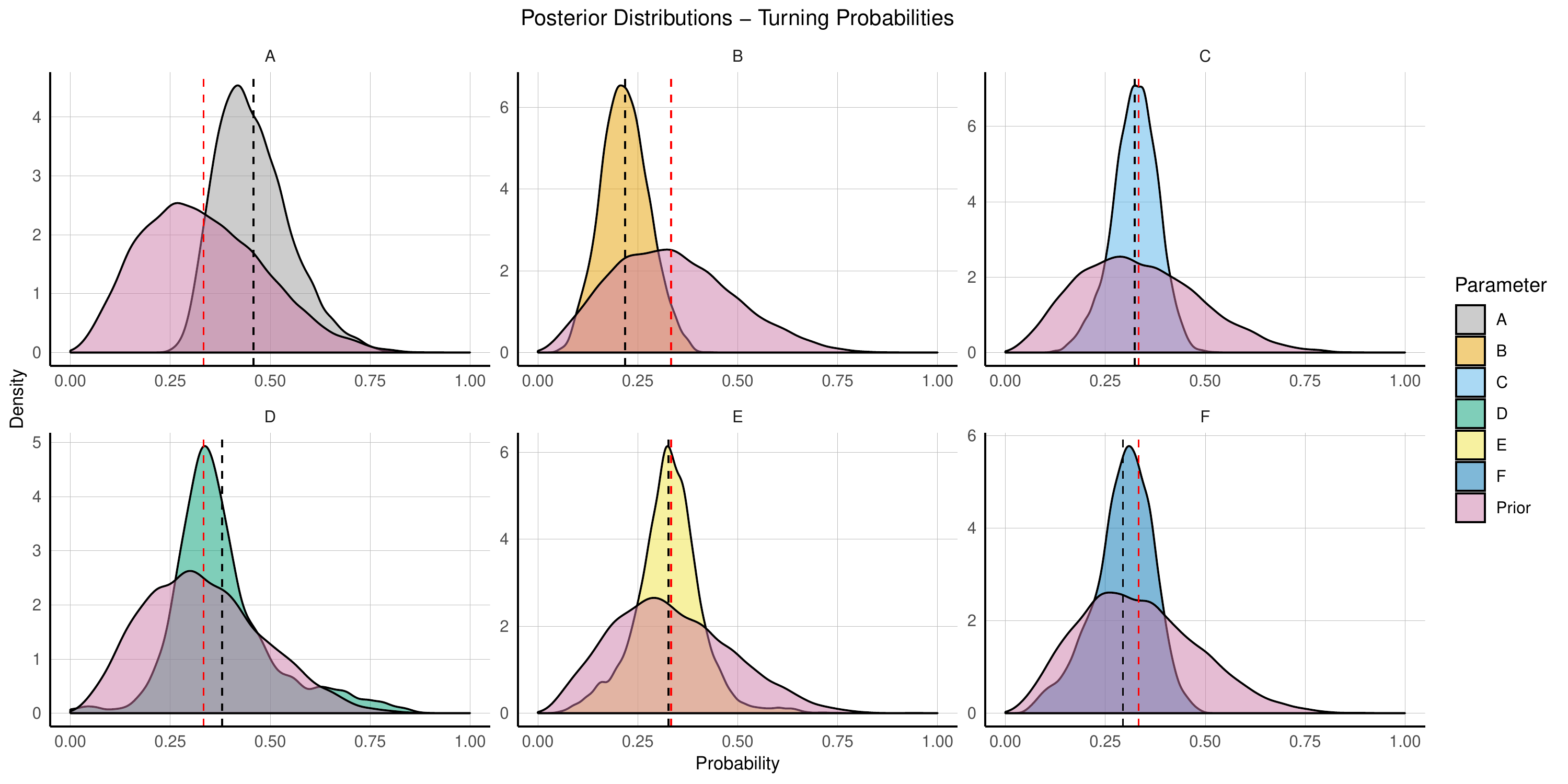}
    \caption{Posterior distributions of each turning probability. The posterior mean is denoted by a dashed black line. The symmetric prior centered on $\frac{1}{3}$ is shown in pink.}
\end{figure}

\textbf{Credible Intervals}
\begin{table}[H]
\centering
\begin{tabular}{@{\extracolsep{4pt}}ccc}
\toprule   
 A & B & C \\ 
\midrule
{[}0.315,0.665{]} & [0.105,0.339] & [0.201,0.4300] \\ 
\bottomrule
\end{tabular}
\caption{95\% Credible Intervals for the parameters at intersection 4} 
\end{table}

\begin{table}[H]
\centering
\vspace{-0.5cm}
\begin{tabular}{@{\extracolsep{4pt}}cccc}
\toprule   
 D & E & F & U \\ 
\midrule
{[}0.185,0.715{]} & [0.149,0.491] & [0.120,0.421] & [24,31] \\ 
\bottomrule
\end{tabular}
\caption{95\% Credible Intervals for the parameters at intersection 3 and the number of pedestrians} 
\end{table}

\textbf{Prediction Intervals} \\
To calculate a 95\% prediction interval for the observed counts, 1000 samples of each parameter were taken from the posterior distributions. These parameters were used to run 1000 simulations of the pedestrian model in NetLogo. The counts from each simulation were used to calculate the 95\% prediction intervals below. \\

\begin{table}[H]
\centering
\vspace{-0.5cm}
\begin{tabular}{@{\extracolsep{4pt}}cccc}
\toprule   
 AW & TR & TA & CPS \\ 
\midrule
{[}234,388{]} & [143,299] & [142,287] & [304,472] \\ 
\bottomrule
\end{tabular}
\caption{95\% Prediction Intervals for the simulated counts at the four entry/exit nodes} 
\end{table}

The algorithm was run for a total of 282 minutes, with a final posterior distribution obtained of 3000 samples at population 10. In the final population, 7,620 simulations were run using sampled probabilities with 39\% of these accepted. The two probabilities A and B in particular have diverged significantly from the prior. A is the probability of turning towards Anglesea Ward, which has a higher count than the two counters in the direction of B. We see that A has a posterior mean of $\mu(A) = 0.488$, in comparison to $\mu(B) = 0.202$. The other probabilities have not shifted significantly from the prior mean, although the variance has reduced. \\

The mean of the simulated counts vs the observed counts are given in table 0.14. They are relatively close, although further improvements can be made by regression adjustment techniques or re-running the model with a modified $\epsilon$ reduction threshold. Further such post analysis is given in the Appendix. Each 95\% prediction interval of the simulated counts in Table 0.17 cover the true observed count. The 95\% credible intervals of the parameters (Table 0.16) are relatively narrow for all parameters except $D$. This is the probability of turning towards the Centre Place intersection from Ward/Worley. The posterior distribution of $D$ appears to be multi-modal. At higher probabilities there is a greater likelihood than that of the prior. \\

\section{Discussion and future work}
The pedestrian model described above serves as a demonstration of using ABC-SMC to validate and infer parameters of a pedestrian ABM. While pedestrian modelling is a common tool used by local government agencies to understand how people use walking as a mode of transportation, there is a challenge in creating the model as an accurate representation. ABC-SMC was found to be an effective tool for validating the simplified Hamilton CBD model such that simulated data closely matches the observed data. Once a validated model has been obtained, the intricacies of the real world system can be viewed through parameter inference. ABC-SMC is also effective for this purpose. It was able to find a precise posterior distribution of probability parameters for turning in each possible direction at the two intersections in the model. \\

As seen in Figure 0.22 of the Appendix, the posterior mean converges to a stable state for some parameters by $\sim$ Population 7, while others have not converged. In addition, the variance is continuously decreasing up to Population 10. This suggests that running the model for more populations will give a more precise posterior. However this will come with further computation time. A balance must be obtained between the precision of the posterior and computation time. Such a balance is a common thread in statistical models such as ABC. \\

While the model ran in a relatively short amount of time, extending the network to include more intersections and observed data will result in an increase in computation time. There remains further research to be conducted in the areas of optimal kernel choice, epsilon threshold, and prior choice. All three significantly affect the computation time. Another option is to increase computing resources. The sequential nature of ABC-SMC does not lend itself well to parallelisation in comparison to other techniques such as ABC-MCMC. However, within populations there is an opportunity to efficiently distribute both the particle pertubation and simulations across a cluster computing network. This was not investigated in the Hamilton CBD model, but will be necessary when extending the model to a larger area of the CBD. \\

Further improvements could be gained using dimensionality reduction methods. If the model is extended to a larger area of the CBD, the dimensionality of the observed data will increase, hence reducing the likelihood of simulating data sufficiently close to the observed data. Methods such as \textit{random projections} could be investigated to reduce the dimensionality of the data and decrease computation time. \\

A vital aspect of developing a useful pedestrian model is in the breadth of data collected in the network. Currently there is limited data collected on pedestrian counts. Ideally, counts would be recorded at all nodes of the CBD network. This would enable the creation of a more realistic model which is able to explain pedestrian flows. The use of the mock data in the Hamilton CBD model means that the parameters inferred from the model are not necessarily the same as the hidden parameters in the real world complex system. HCC is investigating more widespread data collection of pedestrian counts in the CBD. Once this data is available, the extended model can be used to accurately answer questions about pedestrian flows in the Hamilton CBD. \\

While the Hamilton CBD model focuses solely on pedestrian flows, other transportation flows could also be modelled. ABM is well suited to simulating any type of agent based movement through a network. One such example is modelling the traffic flow of vehicles at an intersection, with parameters such as probabilities of turning or average speed through the intersection. Following the development of an ABM to enable this simulation, ABC-SMC can be applied to validate and infer parameters of the model. \\

\newpage
\section{Appendix}

\begin{figure}[H]
  \centering
    \includegraphics[width=16cm]{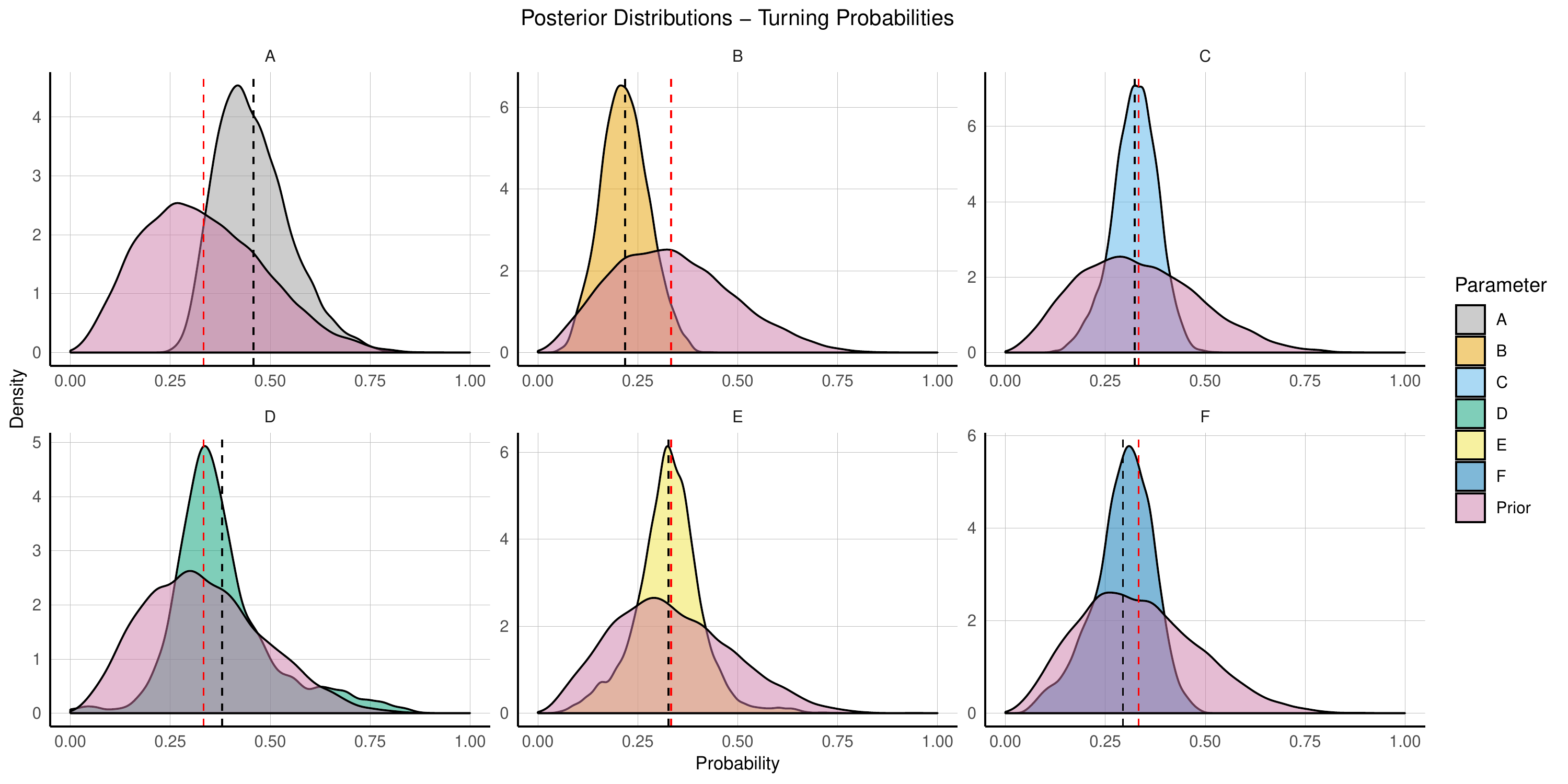}
    \caption{Posterior distributions of each turning probability with the prior distribution shown in pink as a symmetric Dirichlet distribution.}
\end{figure}
\vspace{-0.5cm}
The prior mean of $\frac{1}{3}$ is shown in red, while the posterior mean of each parameter is shown in black.

\vspace{0.5cm}
\begin{figure}[H]
  \centering
    \includegraphics[width=16cm]{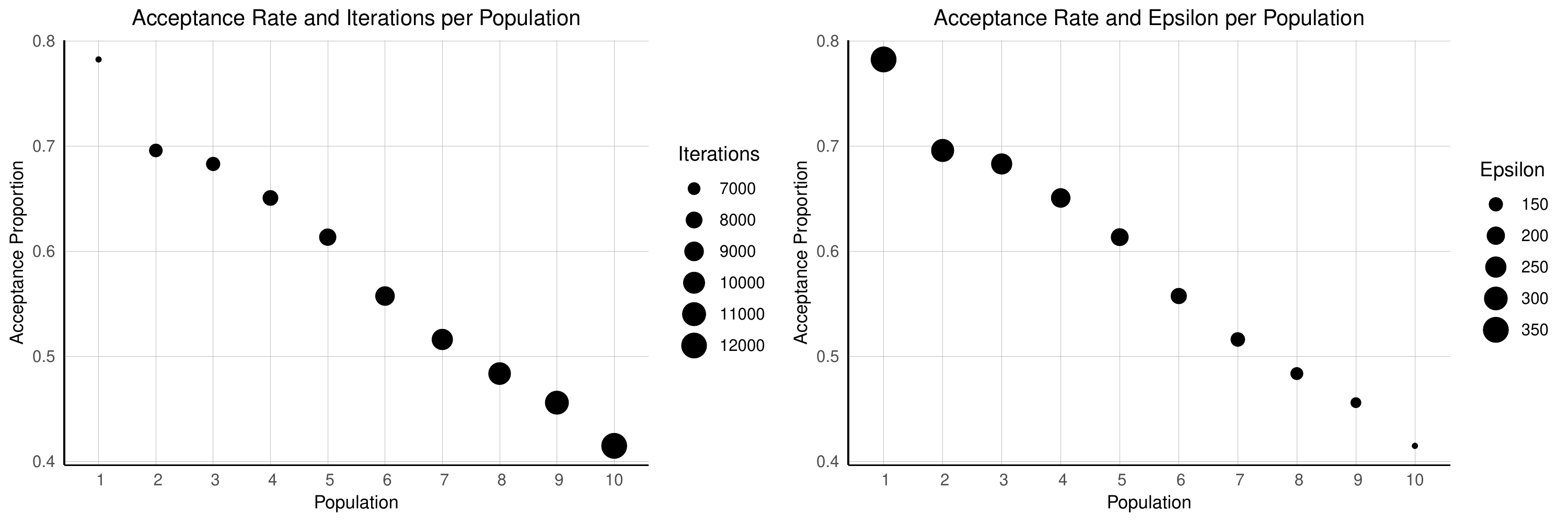}
    \caption{Left: The change in acceptance rate and number of iterations per population. Right: The change in acceptance rate and epsilon per population.}
\end{figure}
As seen in the figure above, the ABC-SMC algorithm maintains a relatively high acceptance rate of $\sim$40\% at the final population. Relative to the reduction in epsilon, a large number of samples are accepted. If the same epsilon thresholds were used in the standard Rejection ABC algorithm, due to the continued sampling from the prior the acceptance rate would typically be far lower.

\begin{figure}[H]
  \centering
    \includegraphics[width=16cm]{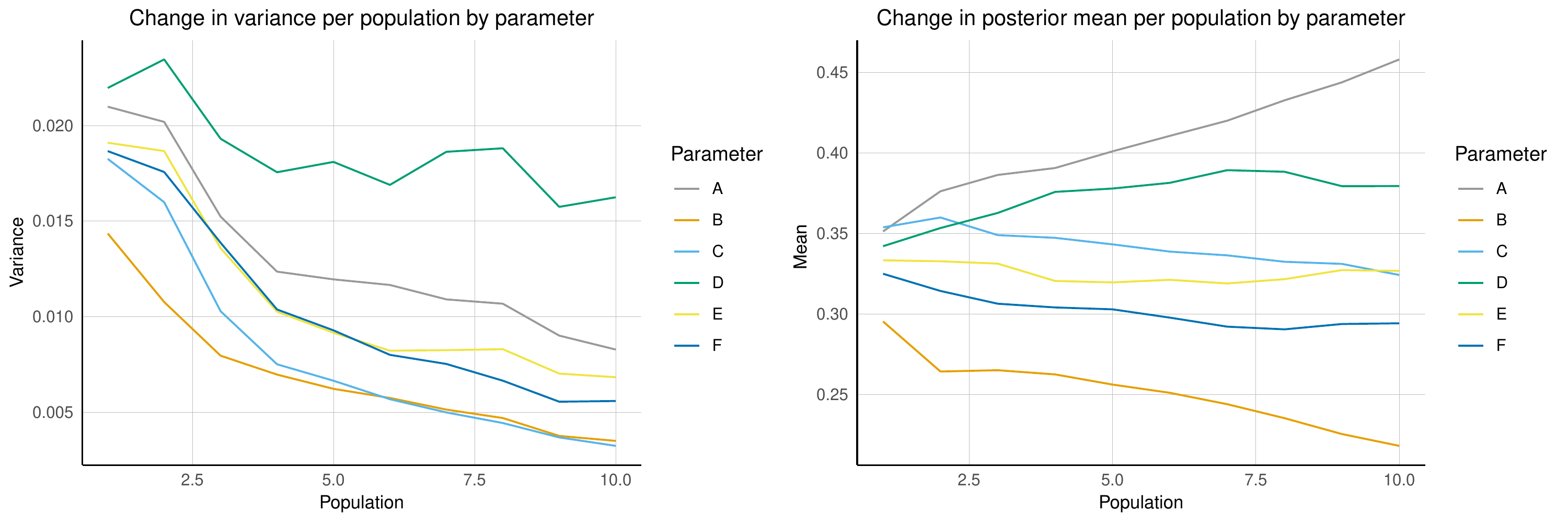}
    \caption{Left: The change in variance per population by parameter. Right: The change in the posterior mean per population by parameter.}
\end{figure}

\textbf{Regression Adjustment}
\begin{figure}[H]
  \centering
    \includegraphics[width=16cm]{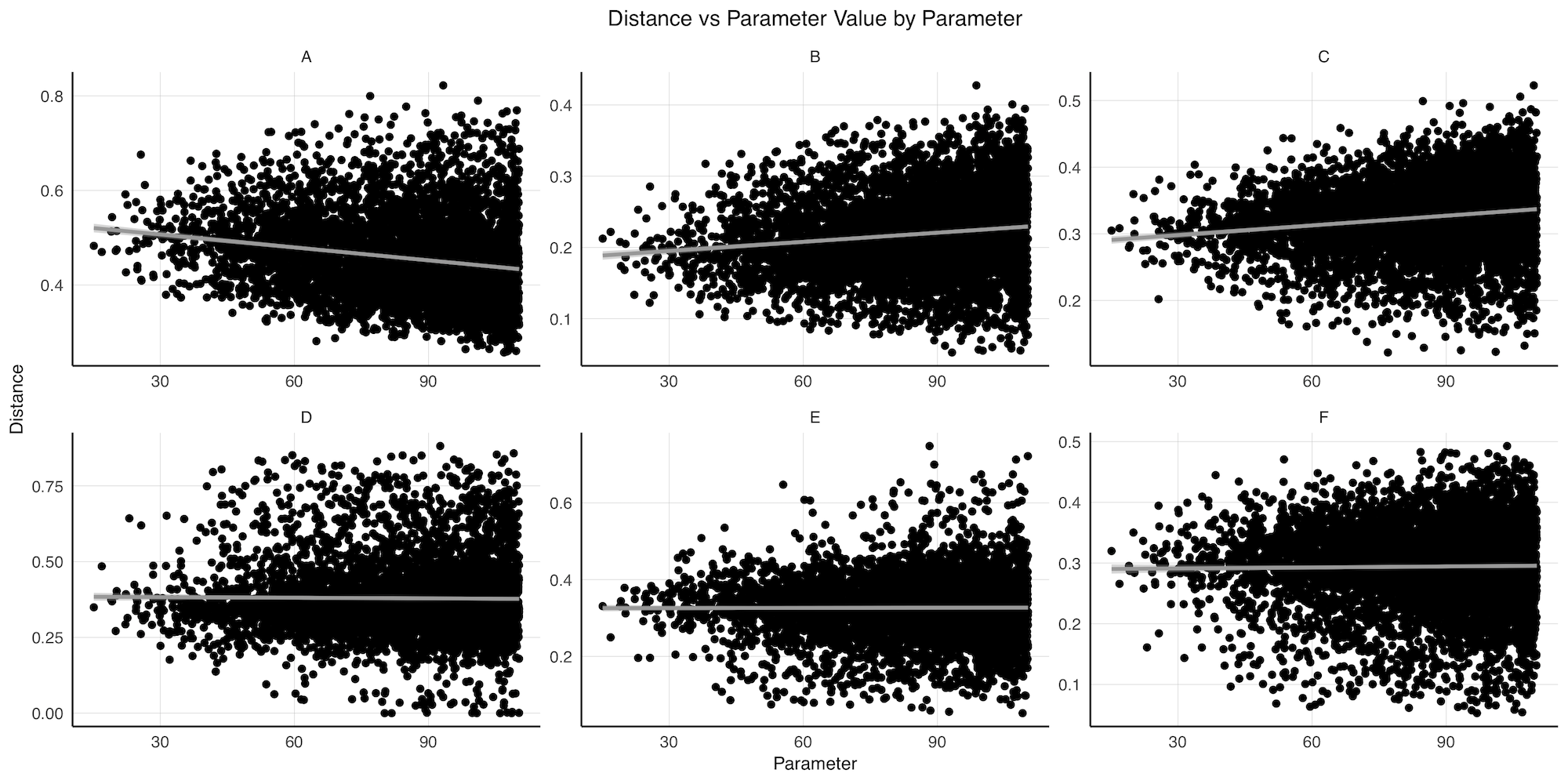}
    \caption{The relationship between the distance measures and value of each parameter.}
\end{figure}

As seen in the above plot, the parameter values seem to be reducing in variance as the distance measure decreases. Theoretically, as the distance $d \rightarrow 0$, the algorithm will converge to the true parameter values $\theta$. As described in Section 5.4.1, this relationship can be investigated to further improve the accuracy of the posterior distributions. Below, the value of each $\theta$ is calculated using a linear model predicting the parameter value using the distance measure:
\begin{align}
\text{Parameter} = \beta_{0} + \beta_{1} \times \text{Distance}
\end{align}
In reality, this relationship is non-linear and the R-squared is extremely low, but the model serves only as a method to find a more precise parameter and not predict a parameter based on distance. The adjusted posterior means are calculated as:

\begin{table}[H]
\centering
\begin{tabular}{@{\extracolsep{4pt}}llccccccc}
\toprule   
 A & B & C & D & E & F \\
\midrule
0.520 & 0.189 & 0.291 & 0.384 & 0.326 & 0.290 \\
\bottomrule
\end{tabular}
\caption{Regression Adjusted Posterior Means} 
\end{table}

Simulating using these posterior means, we obtain the following simulated counts:

\begin{table}[H]
\centering
\begin{tabular}{@{\extracolsep{4pt}}llll}
\toprule   
 Counter & Observed Count & Simulated Count & Difference \\ 
\midrule
AW & 321 & 335 & 14\\ 
TR & 222 & 222.9 & 0.9\\ 
TA & 202 & 202.5 & 0.5\\ 
CPS & 380 & 393.5 & 13.5\\ 
\bottomrule
\end{tabular}
\caption{Observed vs simulated counts following regression adjustment} 
\end{table}

The differences are improved significantly for the counts at TR and TA, but do not change significantly for the counts at AW and CPS.

\end{document}